\documentclass[pdflatex,sn-mathphys-num]{sn-jnl}% Math and Physical Sciences Numbered Reference Style
\usepackage{graphicx}%
\usepackage{multirow}%
\usepackage{amsmath,amssymb,amsfonts}%
\usepackage{amsthm}%
\usepackage{mathrsfs}%
\usepackage[title]{appendix}%
\usepackage{xcolor}%
\usepackage{textcomp}%
\usepackage{manyfoot}%
\usepackage{booktabs}%
\usepackage{algorithm}%
\usepackage{algorithmicx}%
\usepackage{algpseudocode}%
\usepackage{listings}%
\usepackage{bbm}
\usepackage{multirow}

\usepackage{algorithm}
\usepackage{algpseudocode}
\usepackage{array}
\usepackage{graphicx}
\usepackage{textcomp}
\usepackage{listings}
\usepackage{xcolor}
\usepackage{amsmath,amssymb,amsfonts}
\definecolor{codegreen}{rgb}{0,0.6,0}
\definecolor{codegray}{rgb}{0.5,0.5,0.5}
\definecolor{codepurple}{rgb}{0.58,0,0.82}
\definecolor{backcolour}{rgb}{0.95,0.95,0.92}

\lstdefinestyle{mystyle}{
    backgroundcolor=\color{backcolour},   
    commentstyle=\color{codegreen},
    keywordstyle=\color{magenta},
    numberstyle=\tiny\color{codegray},
    stringstyle=\color{codepurple},
    basicstyle=\ttfamily\scriptsize,
    breakatwhitespace=false,         
    breaklines=true,                 
    captionpos=b,                    
    keepspaces=true,                 
    numbers=left,                    
    numbersep=5pt,                  
    showspaces=false,                
    showstringspaces=false,
    showtabs=false,                  
    tabsize=2
}

\newcolumntype{C}[1]{>{\centering\arraybackslash}p{#1}}
\theoremstyle{thmstyleone}%
\theoremstyle{thmstyletwo}%

\theoremstyle{thmstylethree}%

\begin{document}

\title[Leveraging LLMs for reward function design in reinforcement
learning control tasks]{Leveraging LLMs for reward function design in reinforcement
learning control tasks}


\author*[1]{\fnm{Franklin} \sur{Cardenoso}}\email{fracarfer5@gmail.com}

\author[1]{\fnm{Wouter} \sur{Caarls}}\email{wouter@puc-rio.br}


\affil*[1]{\orgdiv{Departament of Electrical Engineering}, \orgname{Pontificial Catholic University of Rio de Janeiro}, \orgaddress{\street{Rua Marquês de São Vicente, 225}, \city{Rio de Janeiro}, \postcode{22451-900}, \state{RJ}, \country{Brazil}}}

% \affil[2]{\orgdiv{Department}, \orgname{Organization}, \orgaddress{\street{Street}, \city{City}, \postcode{10587}, \state{State}, \country{Country}}}

% \affil[3]{\orgdiv{Department}, \orgname{Organization}, \orgaddress{\street{Street}, \city{City}, \postcode{610101}, \state{State}, \country{Country}}}


% WOUTER: IN THE ENTIRE DOCUMENT, LIMIT USE OF WORDS SUCH AS "CRITICAL", "SIGNIFICANT", "BROAD", etc., which make people think the text was generated by an LLM.

\abstract{The challenge of designing effective reward functions in reinforcement learning (RL) represents a significant bottleneck, often requiring extensive human expertise and being time-consuming. Previous work and recent advancements in large language models (LLMs) have demonstrated their potential for automating the generation of reward functions. However, existing methodologies often require preliminary evaluation metrics, human-engineered feedback for the refinement process, or the use of environmental source code as context. To address these limitations, this paper introduces LEARN-Opt (LLM-based Evaluator and Analyzer for Reward functioN Optimization). This LLM-based, fully autonomous, and model-agnostic framework eliminates the need for preliminary metrics and environmental source code as context to generate, execute, and evaluate reward function candidates from textual descriptions of systems and task objectives.
LEARN-Opt's main contribution lies in its ability to autonomously derive performance metrics directly from the system description and the task objective, enabling unsupervised evaluation and selection of reward functions. Our experiments indicate that LEARN-Opt achieves performance comparable to or better to that of state-of-the-art methods, such as EUREKA, while requiring less prior knowledge. We find that automated reward design is a high-variance problem, where the average-case candidate fails, requiring a multi-run approach to find the best candidates. Finally, we show that LEARN-Opt can unlock the potential of low-cost LLMs to find high-performing candidates that are comparable to, or even better than, those of larger models. This demonstrated performance affirms its potential to generate high-quality reward functions without requiring any preliminary human-defined metrics, thereby reducing engineering overhead and enhancing generalizability.
}

\keywords{reinforcement learning, large language models, reward engineering, reward function}



\maketitle

\section{Introduction}\label{sec1}

% \wc{Why British English?}

Reinforcement learning (RL), a trial-and-error-based policy optimization approach~\cite{sutton2018reinforcement}, has demonstrated to be a powerful paradigm, achieving remarkable success in a wide range of tasks, from mastering intricate games to advanced robotic control~\cite{shao2019survey,singh2022reinforcement}. However, its great success across diverse domains is highly related to the quality of the reward functions, since well-designed reward signals are essential for guiding the agent's learning process towards desired behaviors and providing the necessary feedback for policy optimization and convergence~\cite{dewey2014reinforcement}.

Given its fundamental importance, designing a practical reward function becomes a challenging aspect of RL development, particularly for complex or high-dimensional tasks. Although this process can be done through reward engineering or reward shaping techniques, the manual design of reward functions is a highly non-trivial process that often requires extensive domain expertise. In fact, quantifying desired outcomes is inherently tricky, making it a time-consuming trial-and-error process~\cite{eschmann2021reward}.

This iterative process can lead to suboptimal behaviors or, in some cases, unintended consequences, as the agent may exploit loopholes in the reward structure rather than achieving the true underlying objective~\cite{ibrahim2024rfdesign}. 
Consequently, the combined complexity and effort required for human-crafted rewards create a major bottleneck, limitating the applicability and scalability of RL systems in more complex scenarios. Therefore, there is a need for automated solutions in reward design to advance the adoption and scalability of these systems.

On the other hand, recent breakthroughs in large language models (LLMs) \cite{zhao2023survey,minaee2024large} have opened new avenues for automating various tasks, including, for instance, high-level decision-making and code generation, which are used in general applications and, more specifically, in robotics and RL~\cite{zhou2024large,10766898}.

More specifically, the LLMs, with their advanced understanding of natural language and specialized coding tasks, offer a promising path to decrease the manual effort and become a powerful tool for automating the reward design process. One of the most representative examples of this paradigm is EUREKA, which demonstrates how LLMs can be leveraged for reward function code generation in an evolutionary scheme~\cite{ma2023eureka}. By using raw environment source code as input, EUREKA performs evolutionary optimization over candidates guided by feedback, achieving impressive performance across different tasks without requiring task-specific prompts. Besides EUREKA, other interesting LLM-powered approaches are TEXT2REWARD \cite{xietext2reward}, CARD \cite{sun2024large}, Auto MC-Reward \cite{li2024auto}, and L2R \cite{yu2023language}, which also leverage LLMs to generate reward functions in human-readable code. 

% Actual limitations
However, despite these advancements, a critical limitation persists across current methodologies: the reliance on \emph{a priori} measurement systems to evaluate candidates. For example, the reward reflection mechanism in EUREKA relies on user-defined metrics, in addition to policy training statistics, for feedback. While the training statistics are inherent to the training process and can be easily calculated, all the effort is then focused on defining the predefined metrics used as a fitness function, which, depending on the task, can also be challenging.  
Similarly, other LLM-based reward engineering frameworks, while reducing human effort in specific aspects, still require human foresight or pre-computation of what constitutes success or progress. For example, by using natural language prompts to define objectives, but requiring active human feedback for refinement processes~\cite{xietext2reward}; or by applying LLMs as a "proxy reward function" based on human-defined criteria~\cite{kwon2023reward}. 

Consequently, while LLMs are used for code synthesis or planning, their application in automated reward design still requires some form of human-engineered preliminary metrics to define the problem space or evaluate the solutions. On the other hand, relying on source code is also unfeasible in some cases, for example when state transitions and reward signals come from a real-world environment. Other LLM-based methods also depend on human knowledge or insights into the task. This dependency limits true automation, especially in novel or ill-defined environments where such metrics are unavailable or difficult to formulate. This underscores the need for automatic metric creation to ensure the robustness and generalizability of the reward design process.

% What we propose
To overcome these limitations, we leverage EUREKA's idea of using LLMs for reward function generation, but address its main limitation of autonomous evaluation. Under these considerations, this paper introduces LEARN-Opt (LLM-based Evaluator and Analyzer for Reward Function Optimization), a framework designed for automatic reward function creation that eliminates the need for any preliminary evaluation signal. LEARN-Opt's core innovation lies in its ability to autonomously design and utilize performance metrics directly from input descriptions in natural language, such as task objectives, system descriptions, and raw numerical execution data, without requiring any prior human definition of success criteria, examples of reward functions, or explicit feature engineering for evaluation. Thereby, exploiting the information concerning the system description and the task objective supplied by the user, expressed in natural language.

The key advantages of the LEARN-Opt framework can be summarized as follows:

\begin{itemize}
    \item LEARN-Opt removes the human effort that involves the manual design of task-specific metrics. This implementation, in addition to the candidate selection mechanism, aims to reduce engineering overhead and improve the scalability of the reward design.
    \item LEARN-Opt enhances applicability to novel or ill-defined tasks where intuition for quantifying success is limited.
    \item The system leverages LLM agents to generate and refine reward function candidates based on autonomously crafted evaluation signals based on the task objective. Thereby, LLMs also act as "judges" of the generated candidates.
    \item By accepting 
    only natural language as input, LEARN-Opt offers a more versatile and applicable input modality, addressing scenarios where the environment's source code may be unavailable or complex for reading and using effectively. 
\end{itemize}

In our experiments, we demonstrate that the LEARN-Opt framework is effective and can find better policies than EUREKA. We also show that our framework successfully unlock the power of low-cost LLMs to find high-performing candidates in a high-variance search space, while eliminating the human-metric bottleneck.

\section{Related work}

Historically, the design of reward functions has been a manual and expert-driven process. This process often involves significant trial-and-error, particularly in tasks where defining the desired behavior is more complex. To mitigate these challenges and accelerate training, reward shaping techniques have been developed to provide more informative feedback~\cite{ng1999policy}. However, while these methods can be effective, they still depend heavily on human intuition and domain knowledge, since poor reward functions can lead to suboptimal behaviors or undesired policies.

In this context, the introduction of LLMs has started a new era in automating reward function generation, leveraging their advanced capabilities in natural language understanding, in-context learning and, more especially, code synthesis, such as NVIDIA's EUREKA framework.  

\subsection{EUREKA: A universal algorithm for automated reward design}

EUREKA represents a notable advancement in automated reward design by leveraging the zero-shot generation, code-writing, and in-context improvement capabilities of state-of-the-art LLMs like GPT-4. Its methodology is built upon three key algorithmic design choices:

\begin{itemize}
    \item \textbf{Environment as context:} Given that LLMs are usually trained on native code, EUREKA leverages LLMs' capability to infer semantic details and identify relevant variables and uses them for the generation of executable Python reward functions in a zero-shot fashion by feeding the raw environment code directly to the LLM.
    \item \textbf{Evolutionary search:} This process involves iteratively proposing batches of reward function candidates and refining the most promising ones to reduce the probability of buggy or ineffective generated reward functions. 
    \item \textbf{Reward reflection:} Automated feedback is constructed by summarizing policy training dynamics in texts, tracking values of individual reward components to detail how the RL algorithm is optimizing these separate components and providing a more intricate diagnosis to generate new and better individuals later. 
\end{itemize}

By combining these three mechanisms, EUREKA has demonstrated human-level performance across a diverse suite of RL environments, outperforming human experts on most of the tasks with respect to the baseline reward functions. Another interesting aspect of EUREKA is the incorporation of human inputs to improve reward quality during the optimization process.

However, a closer examination of the optimization process in EUREKA's reward reflection module reveals that the \textit{fitness function} used to provide performance information, in addition to policy training statistics, implicitly depends on pre-existing or pre-defined performance metrics to guide its evolutionary search. This highlights a serious limitation: although EUREKA automates the refinement of reward functions, it cannot autonomously define what "good" performance is. This reliance on a human-engineered "oracle" for evaluation is the key limitation LEARN-Opt aims to tackle. 

\subsection{Other foundation model-based approaches}

Beyond EUREKA, various other approaches have explored the use of Foundation Models (FMs) to design reward functions in RL. For example, Kwon et al. demonstrated that LLMs can be used as proxy reward functions within the RL framework. By analyzing states with desired behaviors specified in prompts, LLMs can provide reward signals to guide an agent's learning~\cite{kwon2023reward}. Similarly, Rocamonde et al. used a VLM for vision-based RL-tasks where the reward value is calculated as the cosine similarity between the CLIP embedding of the current visual environment state and the embedding of the desired objective expressed in natural language \cite{rocamonde2023vision}. In this way, the agent receives higher reward values when approaching the desired objective.

Subsequently, other works leveraged the code generation capabilities of FMs to create executable reward functions. TEXT2REWARD generates reward function code based on user input, environmental source code and continuous human feedback, supporting zero-shot and few-shot reward shaping \cite{xietext2reward}. Similarly, CARD (Coder and Evaluator for Automated Reward Design) uses a feedback loop for refining reward functions where a \textit{Coder} generates reward code and an \textit{Evaluator} analyzes trajectories to improve function efficiency~\cite{sun2024large}. Auto MC-Reward automates the design and refinement of dense reward functions with a three-part system: a \textit{Reward Designer} that generates Python reward functions, a \textit{Reward Critic} that verifies and refines them, and the \textit{Trajectory Analyzer}, which uses agent interaction trajectories to provide feedback for further refinement \cite{li2024auto}. Additionally, L2R~\cite{yu2023language} translates natural language descriptions of robot motions into reward function code, employing a \textit{Reward Translator} to interpret goals and a \textit{Motion Controller} to optimize actions. VLMs are also used for this purpose; VLM-Car, for instance, generates Python code for sequential sub-tasks to achieve a desired objective based on the initial and goal frames. Once generated, these sub-task programs are verified and corrected before being used as dense reward functions within a traditional RL framework \cite{venuto2024code}. 

A key observation across many of these code-generation approaches is that they require well-structured input, such as environmental code (e.g., Python classes, methods, or variables) to guide the FM. While this can lead to functional code, it can also create challenges if the input code suffers from poor design or is simply unavailable. This is particularly relevant for approaches like TEXT2REWARD \cite{xietext2reward}, EUREKA \cite{ma2023eureka}, and CARD \cite{sun2024large}. Additionally, it is worth noting that while most of these works rely on commercial models like ChatGPT or GPT-4, only TEXT2REWARD has reported an analysis using open-source LLMs such as Llama 2 and CodeLlama.

\begin{figure}[h]
    \centering
    \includegraphics[width=0.85\linewidth]{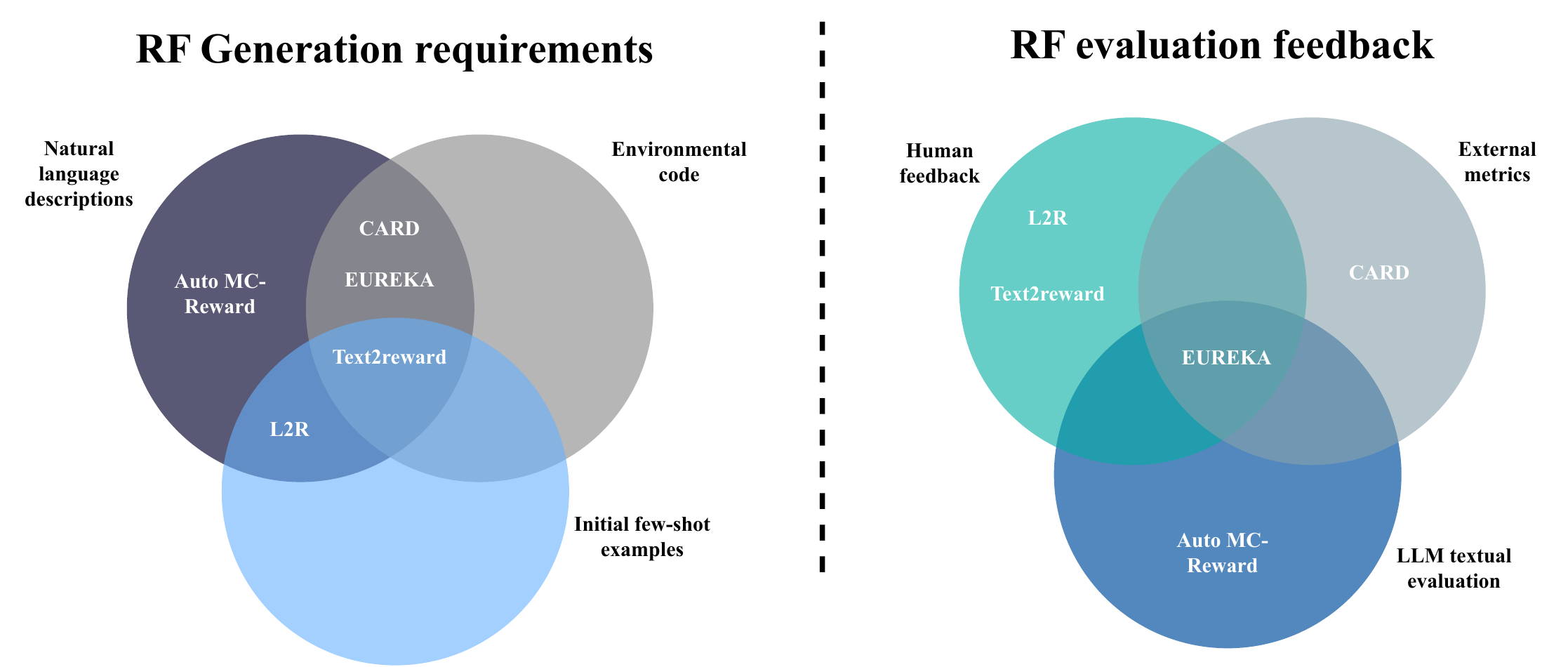}
    \caption{Main related works classification: By generation requirements (left). By evaluation feedback (right).}
    \label{fig:related-works evaluation}
\end{figure}

On the other hand, FMs have also been integrated into Preference-based Reinforcement Learning (PbRL), a framework that models reward functions based on trajectory preferences. As presented by Shen et al. in \cite{shen2024beyond}, LLM4PG is a framework that uses LLMs to automate the human effort in PbRL for trajectory selection in a two-step scheme. It first uses LLMs to abstract numerical trajectory data into natural language descriptions, then it uses LLMs to evaluate pairs of trajectories, generating the preference data needed to build reward models. Similarly, RL-VLM-F uses VLMs with image observations to create the preference labels. Unlike LLM4PG, which abstracts numerical data into natural language, RL-VLM-F directly compares image observations to a text description of the task goal to determine which image aligns better with the task goal. This data is then used to learn the reward model following the standard PbRL framework \cite{wang2024rl}.

While applying FMs within these frameworks can be convenient, other inherent limitations appear, such as the computational cost and the time required for these models to process all incoming information. To address these limitations, Zheng et al. present ONI~\cite{zheng2024online}, an asynchronous architecture designed to decrease reward model and policy optimization time by ranking captions of observations asynchronously with an LLM server and use these rankings within a PbRL framework to learn a reward model more efficiently.

In this context, ORSO tackles limited training budgets by framing the selection of reward functions as an exploration-exploitation problem that evaluates the potential of different reward function candidates and selects the most promising option for refinement based on pre-defined performance metrics~\cite{zhang2024orso}.

Inspired by the success of EUREKA, STRIDE combines environmental source code with a detailed system description within the same evolutionary architecture to reward function generation~\cite{wu2025stride}. Similarly, ROS optimizes the reward function candidate space by creating a "reward observation space" that contains relevant environmental state information that is used to create better and more aligned reward functions to the task objective. At the same time, to evaluate progress, it presents a reconciliation strategy that uses human feedback to define the evaluation metrics \cite{heng2025boosting}.

Finally, ProgressCounts re-frames the reward engineering problem by tasking LLMs with generating "progress functions" instead of tasking LLMs with direct generation of complex and detailed reward functions to estimate how the agent has advanced towards its objective~\cite{sarukkai2024automated}. Although, to assist the LLM, the user provides a high-level description of the task and a small library of helper task-specific functions in the form of programming tools that the LLM can incorporate into its generated code. 

In summary, while the field presents diverse approaches, a critical gap persists. Code-based frameworks are reliant on clean, readable, and available source code. VLM-based frameworks can be computationally expensive and struggle to handle complex, sequential logic. Finally, other methods still require a human-in-the-loop or a pre-defined evaluation mechanism to function. None of these approaches provides a truly autonomous end-to-end framework for reward function design that could operate only from high-level natural language. Recognizing this, LEARN-Opt is designed to fill this methodological gap.

%%%%%%%%%%%%%%%%%%%%%%%%%%%%%%%%%%%%%%%%%%
\section{LEARN-Opt (LLM-based Evaluator and Analyzer for Reward functioN Optimization) framework}

The LEARN-Opt framework is designed as an iterative, closed-loop system for the autonomous generation and refinement of reward functions for RL systems. Its architecture ensures that the entire process operates without requiring any preliminary metrics or posterior human intervention. To accomplish this goal, LEARN-Opt leverages the general knowledge and the coding capabilities of LLMs to first, use natural language information concerning the system description and the task objective to guide the reward generation procedure; and then, generate its own evaluation criteria aligned to the task objective autonomously, inferring what is "good performance" from the given input information, thereby, addressing the limitation presented by previous work.

Our choice to use the user input information expressed in natural language, rather than the raw environmental source code, has multiple advantages. Firstly, natural language provides a more intuitive and accessible interface for human users, including domain experts who may not possess deep programming expertise or familiarity with the source code of a given environment. This decision aims to lower a barrier, allowing LEARN-Opt to capture human intent more directly. Secondly, natural language offers higher flexibility and expressivity when defining complex task objectives and desired behaviors. Finally, relying on natural language mitigates limitations associated with environmental source code. For instance, in many real-world scenarios, the environment's source code may be proprietary, unavailable, or too complex to analyze and extract relevant information. Thus, by accepting natural language descriptions, the system provides a more versatile and applicable input modality.

Furthermore, by combining different LLM agents in the module, using multiple analyzing agents and a multi-run strategy, LEARN-Opt aims to eliminate the human bottleneck and effort associated with manually designing, refining, and validating objective-specific metrics, a process that often demands extensive domain expertise and iterative trial-and-error. %Moreover, it enhances the generalizability and applicability to novel tasks where human intuition for quantifying success is limited or non-existent. In such scenarios, pre-defined metrics are either difficult to formulate or lead to suboptimal results.

As such, since its architecture leverages LLMs to both generate and refine reward function candidates based on its own designed metrics% instead of acting just as judges of preferences
, LEARN-Opt can be considered a self-sufficient system capable of discovering new reward functions autonomously.% within a multi-run strategy.
%\wc{The weakness here is that the metrics are not optimized. Therefore, you're basically saying that an LLM can one-shot a metric better than a human can design it.}

\subsection{Problem formulation}

Let $\mathcal{R}_A$ denote the space of candidate reward functions for an agent $A$. Each $r \in \mathcal{R}_A$ represents a distinct reward function that can be utilized to guide the agent's learning process. Furthermore, we define $\mathcal{L}(\cdot; r)$ as the specific RL algorithm responsible for training an agent using the reward function $r$. It is important to note that $(\cdot)$ accounts for the standard inputs required by the algorithm, such as the environment dynamics, training iterations, and initial conditions. Finally, let $F$ be the fitness function designed to quantitatively evaluate the performance and quality of the agent after it has been trained by $\mathcal{L}$.

Given these definitions, the core challenge addressed by this work is the \textit{reward generation problem} (RGP), where the objective is to identify the optimal reward function $r^*$ from the set $\mathcal{R}_A$ that maximizes the fitness of the resulting trained agent. This optimization problem can be formally expressed as:

$$r^* = \underset{r \in \mathcal{R}_A}{\operatorname{argmax}} \left( F(\mathcal{L}(\cdot; r)) \right)$$

In this context, the origin and nature of the fitness function ($F$) can be derived from user-defined metrics, which are explicitly provided by a human operator depending on the task objective, or determined automatically. For our case, LEARN-Opt automatically derives $F$ from autonomously generated metrics, which are determined from the system description and the task objective without any other preliminary human definition or guidance.

\subsection{Overall description and workflow}

\begin{figure}%[h]
\centering
\includegraphics[width=1\textwidth]{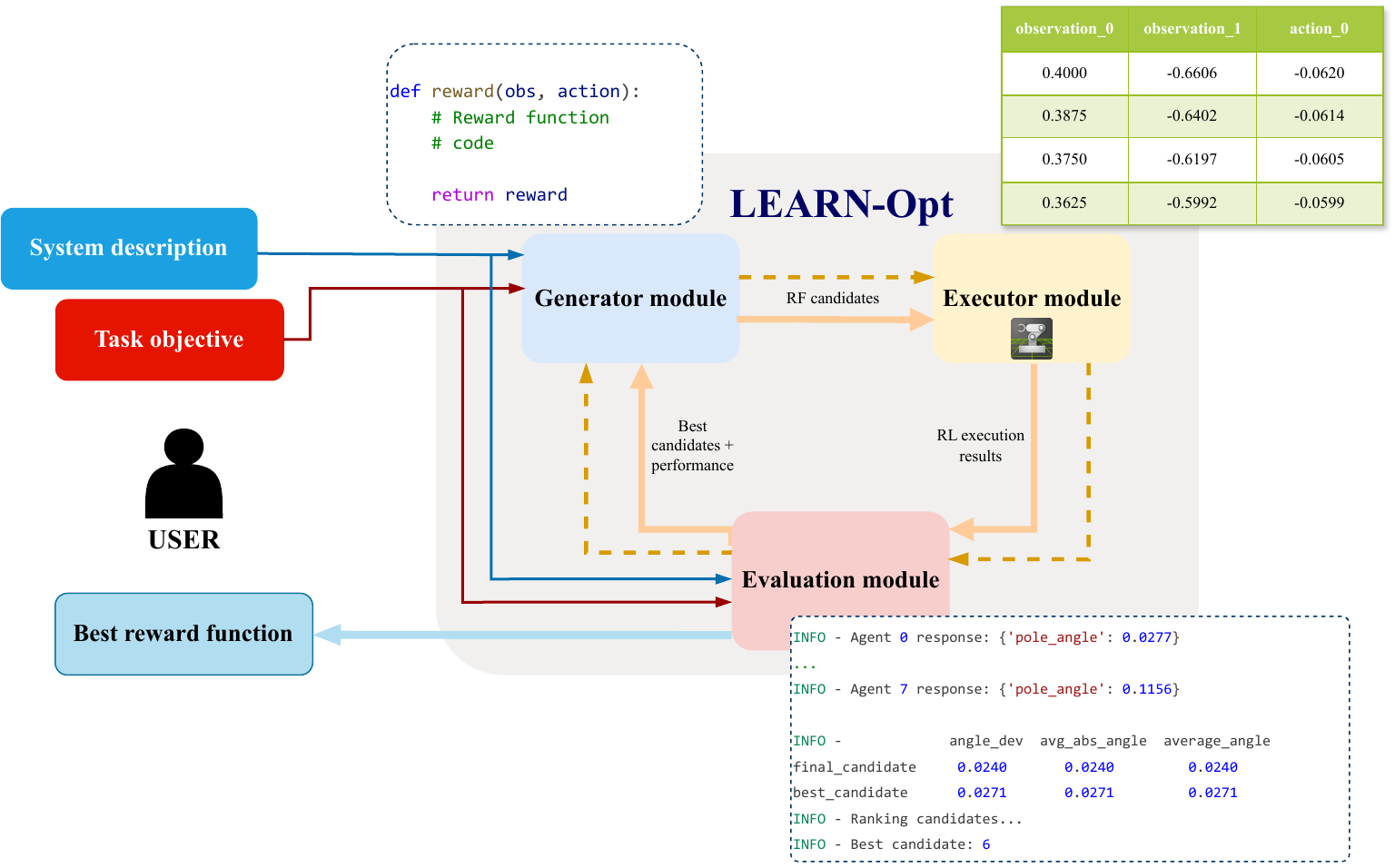}
\caption{Overall workflow of the LEARN-Opt framework: The process begins with the system description and the reward function specification. The generator uses this information to generate candidate reward functions, which are sent to the execution module. After the reward function candidates are tested and their results are collected, these results are passed to the evaluation module, which determines the best reward function candidates in each loop of the optimization process.}\label{overall_workflow}
\end{figure}

The LEARN-Opt framework operates as an iterative, evolutionary, and closed-loop system, drawing inspiration from EUREKA's architecture. We integrate a \textit{generator} module, an \textit{execution} module, and an \textit{evaluation} module to perform the different operations, where the first stage is a \textit{reward engineering} stage, followed by a \textit{reward optimization} stage. Finally, to assess the candidates, LEARN-Opt autonomously creates evaluation metrics. Figure~\ref{overall_workflow} presents the overall workflow of the presented framework.

The high-level workflow proceeds as follows:

\begin{enumerate}
    \item The process begins with a \textit{reward engineering} stage, where the \textbf{generator module} generates a batch of initial reward function candidates based on the system description and the task objective.
    \item These initial candidates are then deployed within the \textbf{execution module} by leveraging an RL pipeline, and raw numerical execution data (state trajectories, actions, observations) is collected and passed to the third module.
    \item The \textbf{evaluation and selection module} takes the generated raw data, the task objective, and system information and autonomously creates performance metrics and a ranking criterion.
    \item Based on these defined metrics, the module evaluates the candidates, ranks them, and selects the best-performing reward function candidate.
    \item An iterative refinement process called \textit{reward optimization} stage begins, where the best candidate and its respective metric values are provided as feedback to the generator module, guiding the generation of improved reward functions in the subsequent iterations. This cycle continues until a certain number of iterations is reached.
\end{enumerate}

\subsection{Module 1: Reward function generation module}

\subsubsection{Initial candidates proposal}

During the reward engineering stage, this module generates the initial batch of reward function candidates by using the natural language input information in the following three-step process. First, an LLM-based mapping agent receives the system description and extracts lists of the states and actions of the system, thereby providing a mapping between vector indices and the names of the state or action dimensions. Next, the generation agent takes the extracted map, the system description, and the task objective, and is asked to generate a batch of \textit{n} distinct reward function candidates in Python code format using a zero-shot (ZS) prompting strategy. Finally, to ensure code quality, we implement an internal evaluation mechanism that serves to assess the generated code by applying a set of basic unit tests (successful execution, output shape/type checking) that ensures that valid code is passed to the next module, so that in case of failing, this mechanism provides feedback to the generation agent to fix the reward function code.

%\wc{I don't understand the mapping agent}
% At the heart of this module, an LLM model powers a mapping agent that first, maps the states and actions of the system given the input description; and a ZS/FS (zero-shot/few-shot) agent that generates batches of diverse reward function candidates in Python code format. Notice that the use of the mapping agent allows for the extraction of the observations and actions of the given system, this information is used to implement an internal fixing mechanism that create unit tests to assess and fix the generated candidates in order to return only valid executable options. This module is fueled by the environment description as context and the task objective, both expressed in natural language and outputs the batch of reward function candidates in code format, and the mapped states and actions. Figure \ref{generation_module} shows the internal implementation of this module.

\subsubsection{Iterative refinement and mutation}
This process corresponds to the reward optimization stage. During this process, instead of generating candidates from scratch, the generation agent mutates the best-performing candidate from previous iterations. This is done by implementing an evolutionary search process in an elitist manner, where the agent, in addition to the input information, receives the code and performance metrics of the best candidate from the analyzer module. Then, the agent applies an in-context reward mutation mechanism using a few-shot (FS) prompting strategy, meaning that the reward generating agent experiences a single conversation across the entire optimization.

Therefore, the reward functions' code of the best previous candidates serve as examples, and their performance metrics results are used as feedback, providing targeted insights into the reward function's performance and areas for improvement. Finally, the agent is tasked with generating \textit{n} variations of the best-performing candidates. This procedure allows the framework to refine and optimize the most promising solutions iteratively.

\begin{figure}%[h]
\centering
\includegraphics[width=0.8\textwidth]{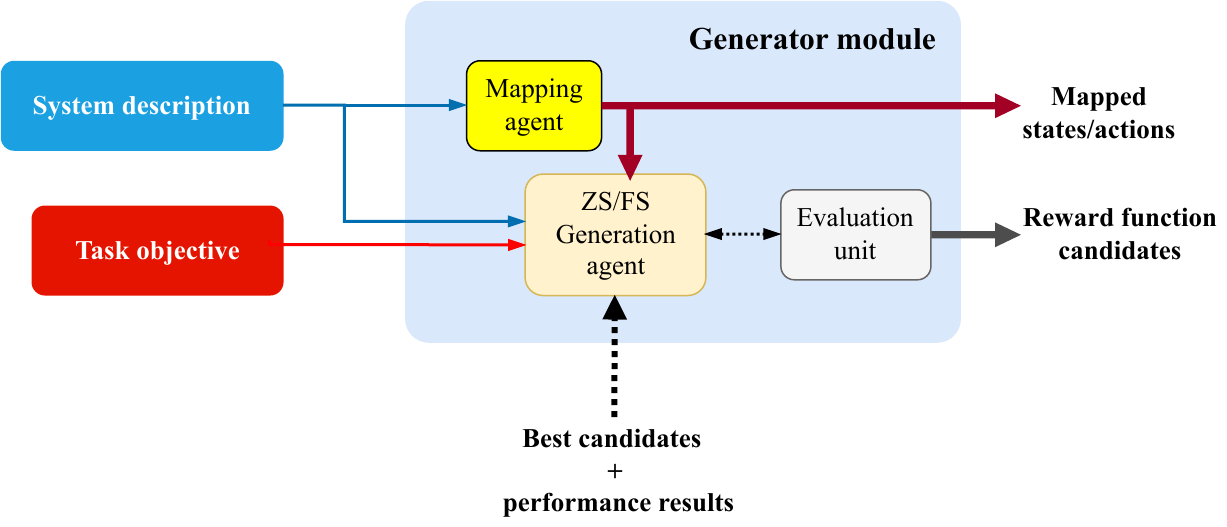}
\caption{Generator module. The mapping agent maps the system description's states and actions and returns this information to the ZS/FS agent to generate valid candidates.}\label{generation_module}
\end{figure}

Figure \ref{generation_module} shows the internal implementation of this module.

%\wc{I'm not sure I would be able to reproduce the implementation based on only this single paragraph}
% This module also implements an evolutionary search process, where the LEARN-Opt framework iteratively propose new batches of reward function candidates and refines the most promising ones in an elitist manner. This process involves an "in-context reward mutation" mechanism, which is performed during the reward optimization stage. There, the LLM receives the best-performing reward function candidates from the previous iterations, along with the numerical feedback from the evaluation module. During this stage, a few-shot prompting strategy is applied to generate the candidates, taking as examples the information of the best candidates from previous iterations, providing targeted insights into the reward function's performance and areas for improvement.% 

\subsection{Module 2: Execution and data collection module}

\begin{figure}%[h]
\centering
\includegraphics[width=0.7\textwidth]{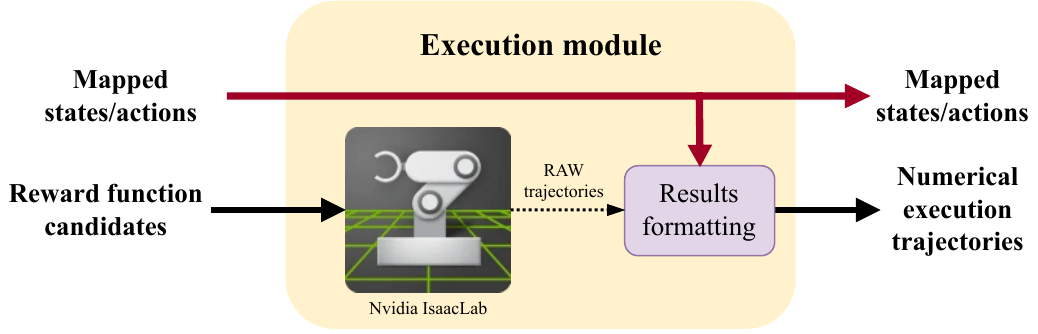}
\caption{Execution module.}\label{execution_module}
\end{figure}

To deploy and test the generated candidates, the execution module integrates NVIDIA's IsaacLab framework as its core (Figure~\ref{execution_module}). Within IsaacLab, an RL policy is trained for each candidate following the traditional RL pipeline. Thereafter, the policy is tested, and the module collects all raw numerical execution data produced by the agent's interactions with the environment, including trajectories composed of executed action sequences and environmental observations. Finally, each raw sequence is formatted in a tabular form, with each column named according to the mapped states and actions.

\subsection{Module 3: Evaluation and selection module}

\begin{figure}%[h]
\centering
\includegraphics[width=0.9\textwidth]{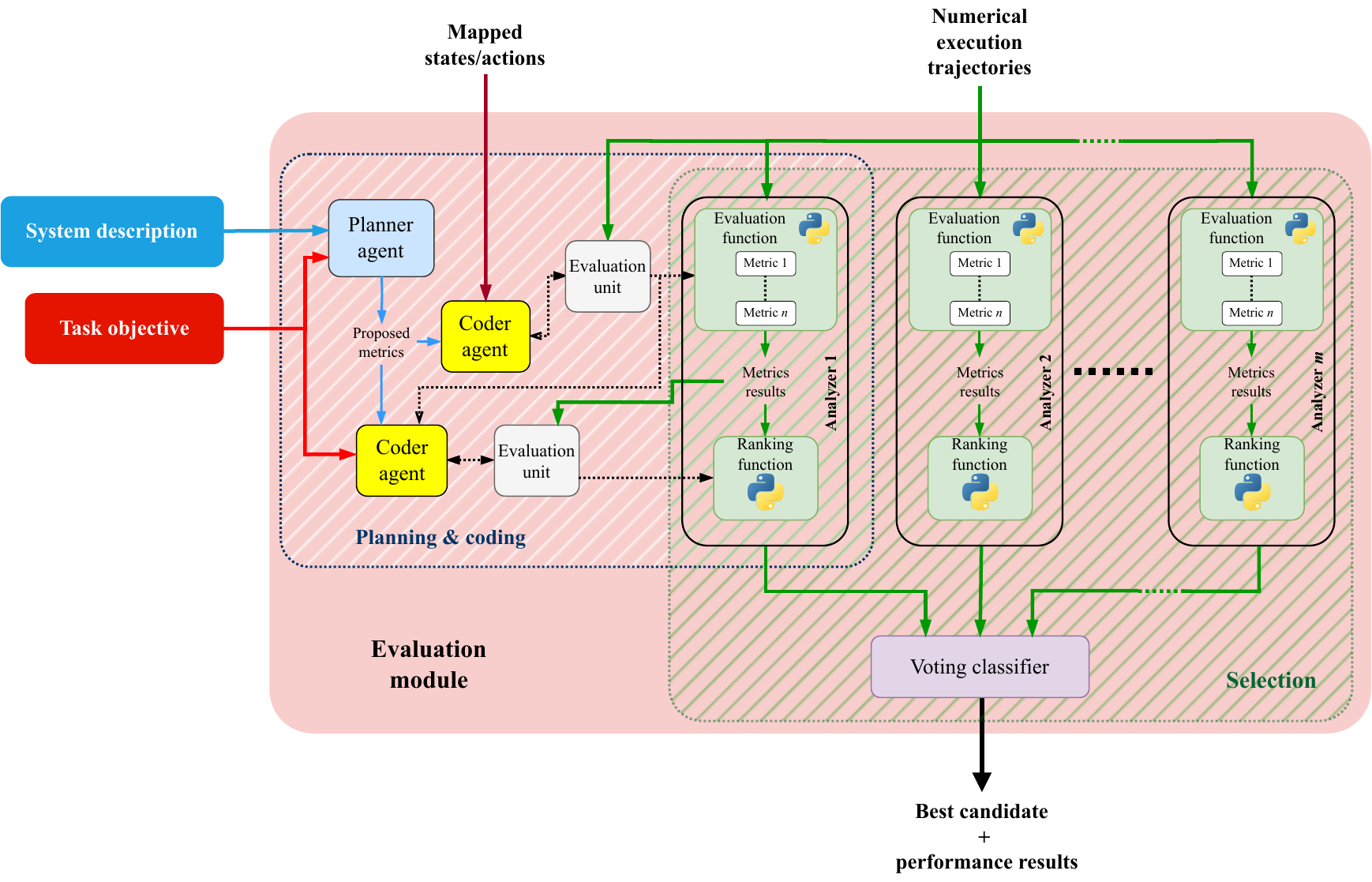}
\caption{Evaluation module. The coder agent is used for coding purposes, while the planner agent is used for a chain-of-thought procedure to generate an evaluation and selection procedure. Its architecture allows the creation of different analyzer sub-modules so that a council can determine the best candidate.}\label{evaluation_module_v2}
\end{figure}

This module autonomously generates evaluation metrics directly from the system description and the task objective, without requiring any human intervention, and then uses these metrics to determine the best reward function from the candidate set. With this, LEARN-Opt aims to achieve autonomy and eliminate the human bottleneck in defining evaluation signals. Figure~\ref{evaluation_module_v2} presents the internal functioning of this module.

Depending on the stage, this module can operate in different modes. First, during the reward engineering stage, it runs a planning and coding process to build a council of \textit{m} analyzer sub-modules. Next, during the reward optimization stage, this module uses the created council to assess and select the best candidate of each iteration loop. Notice that this module is activated just after the execution module finishes executing the reward function candidates and collects their numerical data.

\subsubsection{Council creation: planning and coding}

A key challenge for LEARN-Opt lies in the translation action: how to convert the input information concerning the environment and the task objective into numerical metrics for evaluating the numerical data produced by the execution module and determining the best reward function candidate. Inspired by the success of ensemble architectures in machine learning tasks and the stochastic nature of LLMs, we address this challenge by leveraging the coding capabilities of LLMs to translate the high-level input information into executable Python code and create an ensemble of analysts, a council of \textit{m} different analyzer sub-modules.

To create a single analyzer, we perform a two-step process that we call \textit{planning} and \textit{coding}. 

During planning, we employ an LLM-based planner agent with a zero-shot chain-of-thoughts (ZS-CoT) prompting technique to "think step-by-step" about the task in the following manner:

\begin{enumerate}
    \item First, the planner agent is prompted to extract insights about the desired policy and behavior based on the system and task descriptions.
    \item Next, this reasoning is appended to the context, and the agent is tasked to propose a set of \textit{n} distinct numerical measurements to evaluate the RL agent's performance, their functioning, and the evaluation criteria.
\end{enumerate}

Once the planner agent generates the set of \textit{n} proposed metrics, the coder agent is activated. This agent applies the standard ZS prompting technique to translate the textual descriptions into executable Python code. This is done with the following procedure:

\begin{enumerate}
    \item First, the coder agent receives the generated information of a proposed metric, the mapped list of states and actions, and is tasked to translate the given information into executable Python code to be used as the evaluation function.
    \item To ensure that the agent returns valid code, similar to the generation module, we implement an internal mechanism that samples the numerical data and tests the generated code.
    \item After a valid evaluation function is generated, it is immediately executed on the collected data from the execution module to calculate its numerical metrics results.
    \item This procedure is repeated \textit{n}-times corresponding for each proposed metric.
    \item Next, the coder agent is re-prompted with information about the task objective, the proposed metrics, and the source code of the evaluation function, and tasked with generating a Python function that can sort the candidates' assessments and rank them according to classification criteria; we name it a \textit{ranking function}. This function is also validated through the internal evaluation mechanism.
    \item Then, both functions, the evaluation function and the ranking function, are combined into a single sub-module, which we name the \textit{analyzer sub-module}.
\end{enumerate}

Finally, this two-step process is repeated \textit{m} times to create the council of analyzers.

\subsubsection{Candidate selection: ensemble voting}\label{section:ensemble classifier}

After the council creation is complete and at the end of each iteration, the evaluation module performs candidate selection.
In this mode, the evaluation module receives the numerical data collected by the execution module and sends it to all the \textit{m} analyzers in the council. Then, each analyzer runs its own evaluation and ranking functions and selects its preferred reward function candidate based on its own assessment procedure. Finally, once the set of assessments is complete, the best candidate is determined using a traditional majority voting strategy. By applying this procedure, we ensure that the evaluation module can assess the candidates from different perspectives and select the best candidate based on its unsupervised interpretation.

% Inspired by the success of ensemble architectures in machine learning tasks, we leverage the modular architecture of the evaluation module and its stochasticity to create a more robust decision-making process for the \textit{selection} step. In this way, we can repeat the planning process to create \textit{n} analyzers where each analyzer determines the best candidate according to its own assessment procedure. Finally, once the set of assessments is done, the best candidate is determined by applying a traditional voting strategy. By applying this procedure, we ensure that the evaluation module can assess the candidates from different points of view and select the best candidate based on its unsupervised interpretation, returning the best candidate at the end of the process.

%\wc{The rest of this section lacks structure. Also, suddenly the ZS agent has an internal restart mechanism which has not been explained before!}

\subsubsection{Design rationale}

The intention to modularize the assessment process into different agents and steps offers different advantages: first, the modular architecture is LLM-efficient. Although the council creation can look computationally expensive, its creation occurs only once during the reward engineering stage. During reward optimization, its execution is faster since the module uses already compiled functions. 

Furthermore, by splitting the procedure into distinct, simpler tasks (mapping, generation, planning, and coding), we aim to reduce hallucination, a problem that is frequent in models with small context length or lower capabilities when tasked with complex or too elaborate tasks~\cite{liu2024exploring}.

In addition, implementing the ensemble classifier to determine the best reward function candidate enables a more robust and stable evaluation system. Because of the stochastic nature of LLMs, a single analyzer could center on an ineffective metric. Thereby, by applying a council of analyzers we aim to reduce this variance. Our intuition is that a consensus of different points of view could yield a more robust selection than using a single analyzer.
 
Finally, by implementing a predefined code sanity evaluation mechanism whenever a Python function is generated, we ensure that the module produces functional code, thereby decreasing the failure rate.

% the purpose that each agent can accomplish an specific task, this proposal not only allows a more flexible architecture but also allows to mitigate the hallucination problem that affects LLMs when they are tasked with complex or too elaborated tasks, losing the context or failing to maintain coherence during the generation problem, especially for models with small context length or lower capabilities~\cite{liu2024exploring}. Consequently, this modular architecture allows the integration of different LLM models within a unified system. In addition, implementing the ensemble classifier to determine the best reward function candidate enables a more robust and stable evaluation system, as LLMs can be sensitive to generation likelihood, leading to variance in their outputs. The ensemble analyzer aims to improve the framework's capability to evaluate the different candidates. Another consideration is the ZS agent, which, to avoid undesired behavior, includes an internal restart mechanism that utilizes basic unit tests to generate functional Python code, thereby decreasing the failure rate.

% Another important clarification of the planning and selection steps is that the implementation of the \textit{analyzer} sub-module occurs only during the reward engineering stage. During the reward optimization stage, the framework uses the implemented sub-module directly.

\section{Experimental design}
We evaluate the proposed frameworks under various situations, this section outlines the key guidelines for using both frameworks in these experiments. We begin by describing the baselines, then the LLMs employed, and follow with an overview of the tasks.

\subsection{Baselines}
We use two approaches for our evaluations and compare our results: \textbf{Original}, which is the set of the original reward functions provided by the IsaacLab RL framework by default for each environment, and \textbf{EUREKA}, where the reward function is obtained after running its complete pipeline for the tested environments and tasks.
An important consideration is that since NVIDIA IsaacGym, used in the original paper, was deprecated, the EUREKA version we are using for our comparisons is NVIDIA's released version implemented on top of IsaacLab\footnote{https://github.com/isaac-sim/IsaacLabEureka}.

\subsection{LLM models}
For our experiments, we considered using \texttt{gpt-4.1-nano} because it demonstrated sufficient capabilities for generation tasks, comparable to higher models such as GPT-4.

\subsection{Environments}
We use NVIDIA's IsaacLab 2.2 simulator built on top of IsaacSim 5.4.0 for the environment simulations. This is a framework designed for robot learning and training policies\footnote{https://github.com/isaac-sim/IsaacLab}. For our experiments, we use five environments, which include Cartpole, Quadcopter, Ant, Humanoid, and Franka-cabinet. For the task objectives, we use the same ones as those used by EUREKA; these objectives are presented in Table~\ref{task-objectives}.

Given that there is no official release of the documentation for IsaacLab environments, we adapted the Gymnasium\footnote{https://gymnasium.farama.org/index.html} documentation for the environments we are using to create the system description by combining it with information gathered from the IsaacLab environments' source code. The resulting system descriptions are presented in Appendix~\ref{system_descriptions}.

\begin{table}[h]
\caption{Task objectives}\label{task-objectives}%
\begin{tabular}{@{}cc@{}}
\toprule
\textbf{Task} & \textbf{Objective}  \\
\midrule
Cartpole     & Balance a pole on a cart so that the pole stays upright  \\
Quadcopter   & Make the quadcopter reach and hover near a fixed position  \\
Ant          & Make the ant run forward as fast as possible  \\
Humanoid     & Make the humanoid run forward as fast as possible \\
Franka-cabinet & Open the cabinet door \\
\botrule
\end{tabular}

\end{table}

\subsection{Evaluation metrics}\label{evaluation-metrics}

To perform the optimization process for the EUREKA framework and evaluate our results later, we defined evaluation metrics according to the objective of each task. These metrics will serve to evaluate the validity of the final optimized candidate and validate the optimization process itself. 

\subsubsection{Cartpole}
For the Cartpole task, we use the mean squared error (MSE) to measure the difference between the pole angle and the target angle:
\[
MSE_{\mathrm{angle}} = \frac{1}{n} \sum_{i=1}^{n}(\theta_i - \theta_{\mathrm{target}})^2,
\]
where \(n\) is the number of samples, \(\theta_i\) is the pole's angle, and \(\theta_{\mathrm{target}}\) is the target angle. For this case \(\theta_{\mathrm{target}} = 0\). Better candidates are those which achieve values close to zero.

\subsubsection{Quadcopter}

For the quadcopter task, we use the Euclidean distance between the goal position and actual position of the quadcopter:
\begin{equation}
    d\left( p,q\right)   = \frac{1}{n} {\sum _{i=1}^{n}  \left\lVert q_{i}-p_{i}\right\rVert_2 },
\end{equation}
where $n$ is the number of samples, $q_i = (x_i, y_i, z_i)$ is the actual position of the quadcopter and $p_i = (0, 0, 0)$ is the target position. Similar to the cartpole case, in this task, better candidates are those whose metric values approach zero.

\subsubsection{Ant and humanoid}
For these environments, we use the mean velocity along the \textit{x}-axis (forward direction) as our measurement value:
\begin{equation}
    \bar v_x   = \frac{1}{n} {\sum _{i=1}^{n} v_x },
\end{equation}
where $n$ is the number of samples and $v_x$ is the velocity along the \textit{x}-axis. Contrary to previous cases, better candidates for these tasks are those with higher metric values.

\subsubsection{Franka-Cabinet}

In the Franka-Cabinet environment, we measure the mean number of successful steps when the robot opens the drawer. This is when the drawer opening distance was greater than or equal to 0.35 (derived from the original task description):
\begin{equation}
\bar s  = \frac{1}{n} {\sum _{i=1}^{n}\mathbbm{1}[x_i>=0.35]},
\end{equation}
where $n$ is the number of samples, $x_i$ is the actual drawer position, and $\mathbbm{1}[.]$ represents the indicator function.

\subsection{Experiment details}

To obtain the final reward function candidates, our experimental validation was conducted in two distinct phases: reward function discovery and evaluation. During the reward discovery phase, we ran the entire pipeline for both EUREKA and LEARN-Opt to find an optimized reward function. We repeated the entire procedure 10 times for each framework. For consistency during this phase, all internal RL training processes used a fixed seed, as detailed in Table~\ref{main-parameters}. Notice that although the RL process uses the same seed, the reward generation process is stochastic due to the LLM's temperature, producing different outputs on each run.

In the evaluation phase, each of the ten candidates from both frameworks was independently assessed by training and testing an RL agent five times using the following process. First, we use the same seed as in the discovery phase; this evaluates the candidate's final performance under the same conditions in which the reward function was optimized. Next, four additional RL runs using different unseen seeds are used to evaluate the candidate under different conditions to observe its generalizability.

During testing episodes, we leverage the parallel capabilities of IsaacLab to deploy 128 independent instances with randomized initializations of each agent that simultaneously run the final policy under different conditions. The final policy performance is calculated as the mean of the performance obtained with the metrics calculation across all the parallel environments. As such, a high test performance means the final policy is robust to different initial conditions.

This multi-seed evaluation process allows us to separate the  frameworks' ability to optimize the \emph{policy} from its ability to optimize the \emph{reward function}. If the final goal is to find a good policy, it is not important how the reward function performs on other seeds -- it is simply a means to find a policy. If, on the other hand, we want to find a reward function that generalizes well, the performance on the other seeds is all that matters. We call these two goals the peak performance (PP) and generalization performance (GP).

Finally, we evaluate the frameworks' PP or GP performance as either the average across the 10 runs, or as the maximum. In the first case, we are interested in the expected performance when run once, while in the latter case we want to know how they perform when run 10 times and taking the best result, which combats the high variance of the results. All results are compared against the default reward function.

For the original reward function (IsaacLab's default reward function), we apply the same multi-seed evaluation process as described above. Note that, although LEARN-Opt selects its evaluation measurements during the optimization process, we use the evaluation metrics presented in subsection~\ref{evaluation-metrics} to externally evaluate the performance of the final reward function candidates and therefore validate the results of the LEARN-Opt framework. Note that these metrics were used internally for the EUREKA framework as part of its optimization process.

For all the RL executions, we utilized the A2C algorithm implemented in IsaacLab, with all its RL hyperparameters set to the values detailed in Appendix~\ref{app:ppo_parameters}. We only modified the number of epochs according to each task. Tables~\ref {main-parameters}, \ref{eureka-parameters}, and \ref{learnopt-parameters} summarize the experiment details, and Table~\ref{envs-parameters} summarizes the environment details.

\begin{table}[h]
\caption{Main parameters}\label{main-parameters}%
\begin{tabular}{@{}cc@{}}
\toprule
\textbf{Hyperparameter} & \textbf{Value}  \\
\midrule
Number of experiments per environment (runs) & 10\\
Number of trainings and testing for each final candidate   & 5  \\
Training environments\footnotemark[1] & 4096\\
Testing environments\footnotemark[1] & 128\\
RL algorithm   & A2C  \\
Training seed & 42\\
Testing seeds & 3120, 2190, 6838, 4024\\

\botrule
\end{tabular}
\footnotetext[1]{\textbf{Note:} Since we are using IsaacLab these are parallel vectorized environments.}
\end{table}

\begin{table}[h]
\caption{EUREKA parameters}\label{eureka-parameters}%
\begin{tabular}{@{}cc@{}}
\toprule
\textbf{Hyperparameter} & \textbf{Value}  \\
\midrule
Generations   & 5  \\
Candidates per generation   & 8  \\
Temperature & 1\\
\botrule
\end{tabular}
\end{table}

\begin{table}[h]
\caption{LEARN-Opt parameters}\label{learnopt-parameters}%
\begin{tabular}{@{}cc@{}}
\toprule
\textbf{Hyperparameter} & \textbf{Value}  \\
\midrule
Reward optimization iterations   & 5  \\
Candidates per iteration   & 8  \\
Generator temperature & 1\\
Analyzer ZS-CoT temperature & 0.2\\
Analyzer ZS temperature & 0.2\\
Number of analyzers & 3\\
Number of metrics & 1\\
% Exploitation rate\wc{rate?} & 1\\

\botrule
\end{tabular}
\end{table}

\begin{table}[h!]
\caption{Environment constants}\label{envs-parameters}%
\begin{tabular}{@{}cccccc@{}}
\toprule
\textbf{Parameter} & \textbf{Cartpole}  & \textbf{Quadcopter} & \textbf{Ant} & \textbf{Humanoid} & \textbf{Franka-cabinet}\\
\midrule
Training epochs   & 100   & 100   & 500   & 500   & 1000  \\
Observations        & 4     & 12    & 36    & 75    & 23  \\
Actions             & 1     & 4     & 8     & 24    & 9 \\
Time steps          & 300   & 500   & 900   & 900   & 500\\
\botrule
\end{tabular}
\end{table}

\section{Results and discussion}
This section presents the results obtained from all our experiments\footnote{All the prompts, system descriptions and raw results are available in https://github.com/fracarfer5/learn-opt-appendix}. We will first present the overall results of the proposed framework, then show additional experiments to validate it, and present the corresponding analysis.

\subsection{Overall results}

\begin{figure}
    \centering
    \includegraphics[width=1\linewidth]{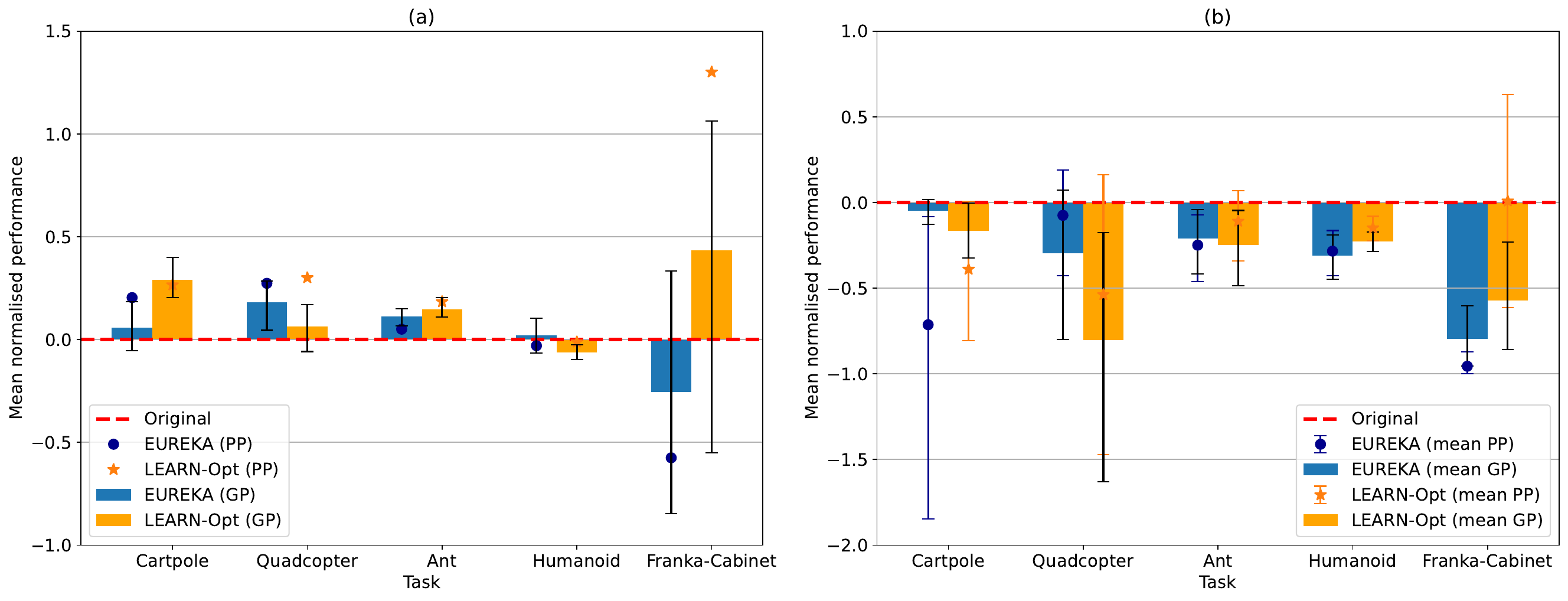}
    \caption{Normalized performance obtained for the evaluated frameworks for: (a) Best results and (b) All results. For both plots, we present the mean and CI values; higher values are better.}
    \label{fig:all-Normalized-metrics}
\end{figure}

Figure~\ref{fig:all-Normalized-metrics} condenses the main results, divided into two parts: First, we analyze the performance of the best candidates across all 10 runs from each framework (Figure~\ref{fig:all-Normalized-metrics}a). Next, we analyze the average performance of the runs to evaluate the reliability and consistency for both frameworks (Figure~\ref{fig:all-Normalized-metrics}b). Given that each task varies in scale, we apply a normalized score (zero value indicated by the dashed red line) based on the four-seed performance of the original reward function to show performance differences across all tasks, ensuring uniform scale. Positive values indicate an improvement over the original reward, and negative values indicate worse performance. In all cases, the error bars represent the 95\% confidence interval (CI).

Figure~\ref{fig:all-Normalized-metrics}a shows the performance of the best candidates from EUREKA and LEARN-Opt for each task for both evaluation criteria: peak performance and generalization. The candidate with the highest generalization performance (GP), which is the mean score from the four unseen seeds, is represented by the bar plot, while the peak performance (PP), which is the score with the original optimization seed, is represented by the markers. Error bars represent the CI over the four seeds.

\textbf{LEARN-Opt is comparable to or better than EUREKA when finding better policies.} 
From this graph, we can observe that LEARN-Opt consistently finds candidates with higher performance and comparable robustness than those found by EUREKA in most cases. For example, for Cartpole, Ant and Franka-cabinet, the LEARN-Opt framework presents a positive GP in Figure~\ref{fig:all-Normalized-metrics}a. While for the Quadcopter and Humanoid environments, EUREKA presents slightly better performance, although the overlapping CI indicate that the results are not statistically distinguishable for most of the cases. 
In addition, the PP values presented for some cases, such as Quadcopter and Franka, indicate that although the generalization performance can be low, the LEARN-Opt framework is able to discover reward functions that exploit the optimization conditions to find policies that outperform considerably better. However, the most interesting result is observed in the Franka-Cabinet environment, where LEARN-Opt clearly shows a greater advantage than the original reward function and EUREKA, which fails almost completely.
Interestingly, in the Humanoid task case, both frameworks struggled to find a reward function that consistently beat the original one, suggesting that this task poses the most challenging reward-design scenario.

On the other hand, Figure~\ref{fig:all-Normalized-metrics}b presents the overall performance, averaging the results across all ten runs obtained by EUREKA and LEARN-Opt, respectively (error bars represent the CI over the ten runs). This plot shows the mean generalization performance (mean GP) and mean peak performance (mean PP). From this plot, we can see that, for all tasks, the mean GP and the mean PP are negative in both frameworks. This reveals that finding a superior, general-purpose reward function is a high-variance problem. However, when contrasting with the previous plot (Figure~\ref{fig:all-Normalized-metrics}a), it suggests that a multi-run approach (ten runs for this case) is necessary to find a high-performance reward function candidate.

In general, from both figures, we can state that to find a policy that could outperform the original baseline, it is necessary a multi-run strategy for both frameworks.
Furthermore, in most tasks, LEARN-Opt achieves better performance than EUREKA, whereas in the remaining tasks, the performance is comparable for both frameworks. Therefore, these results show that LEARN-Opt successfully addresses EUREKA's limitation of relying on predefined user metrics by autonomously creating its own evaluation and selection functions, enabling the framework to find reward function candidates that achieve similar or better performance on the evaluated tasks than state-of-the-art frameworks.

To better support these affirmations, we present Table~\ref{input-ablation-results}, which shows the unnormalized results obtained for the tested environments. These results confirm the generally favorable performance of LEARN-Opt compared to EUREKA and the original reward function, particularly in the PP (policy search) case. 

\subsection{Impact of the metric source vs. optimization backbone} Besides the previous experiments, we conducted an ablation study that decouples the source of evaluation metrics from LEARN-Opt and introduces pre-defined metrics to optimize the generation of reward function candidates. We present \textit{LEARN-opt with metrics}, a variant of our framework that uses the LEARN-Opt backbone but, similar to EUREKA, relies on pre-defined evaluation metrics in the evaluation module to perform the optimization. This allows for a direct comparison of both optimization backbones (\textit{LEARN-Opt with metrics} vs EUREKA), isolating the effect of the optimization procedure itself. Furthermore, by comparing this variant to the full, autonomous LEARN-Opt framework, we can evaluate the impact of our autonomous metric generation. 

\begin{figure}[h]
    \centering
    \includegraphics[width=1\linewidth]{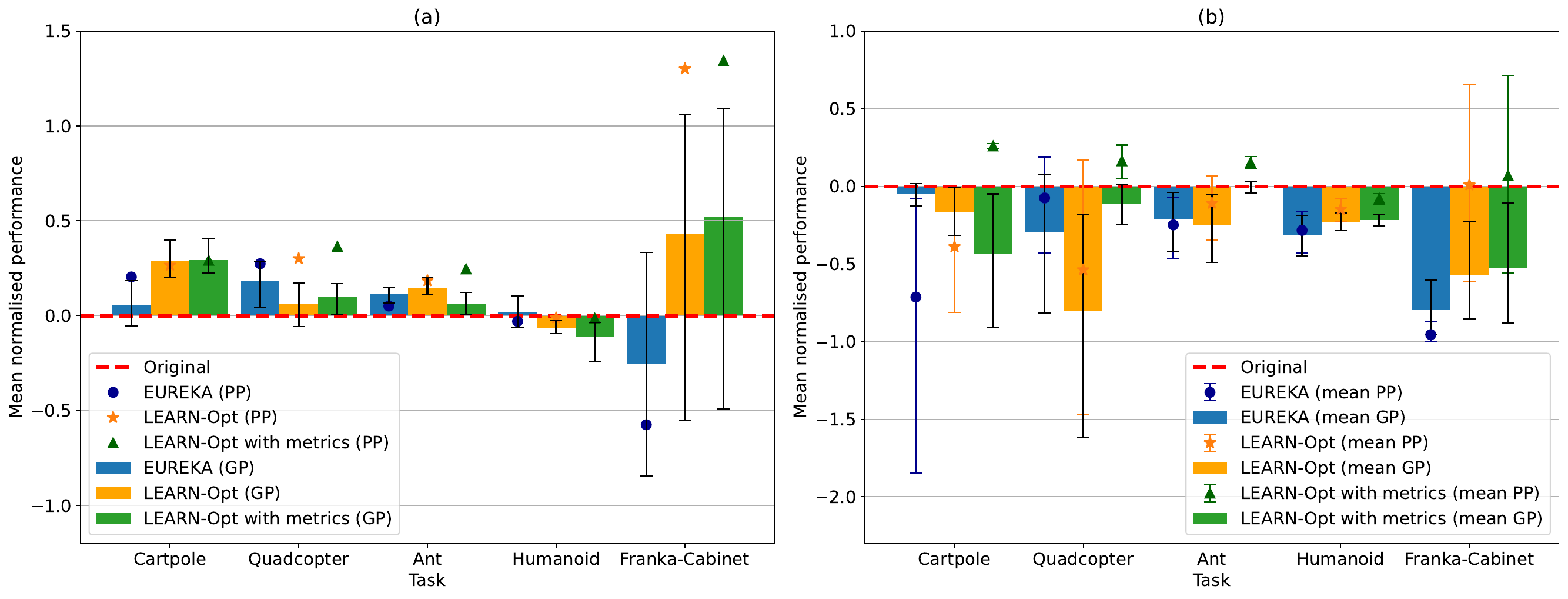}
    \caption{Normalized metrics for EUREKA, LEARN-Opt and LEARN-Opt with metrics: (a) Best results and (b) All results. For both plots, we present the mean and CI values; higher values are better.}
    \label{fig:nlp-ablation}
\end{figure}

The result of this evaluation is presented in Figure~\ref{fig:nlp-ablation} and Table~\ref{input-ablation-results}, which, besides the results presented previously, add the results of \textit{LEARN-Opt with metrics}. From Figure~\ref{fig:nlp-ablation}a, we observe that in general, the LEARN-Opt variant with metrics shows similar peak performance to the proposed framework since the best obtained candidate outperforms EUREKA and the original reward function in most of the tasks. Generalization performance is on par with the original approach.

On the other hand, the most interesting results are presented in Figure~\ref{fig:nlp-ablation}(b), where the introduction of the metrics consistently improves the performance and the variability for the Quadcopter and Ant environments, and slightly improves for the Humanoid and Franka-cabinet environments. Specifically, the PP is now above the baseline for all tasks except the humanoid. These results indicate that LEARN-Opt can be enhanced by providing metrics when available, but is not dependent on them.

After carefully analyzing the input information supplied to both frameworks, we identified that EUREKA's low performance in the Franka-cabinet task is due to the gap between the task objective and the input information provided as environmental source code. This suggests that although the evaluation metrics provide valuable insights to the generator agent during the optimization process, well-defined descriptions about the system and its goal are also necessary to generate better candidates, especially in more complex tasks. This finding also reinforces the fact that the LLM agents are highly dependent on the quality of the input information. 
It is important to notice that in our case, the system descriptions were adapted from the Gymnasium's website and using the information from IsaacLab environments' source code. 

Surprisingly, we observe that EUREKA often produced a lower PP than GP result, while for both LEARN-Opt cases the behavior is reversed. This could indicate that LEARN-Opt produces reward functions that overfit the optimization seed and do not generalize. If the ultimate goal is to find an optimal policy this is not an issue, but it shows room for improvement if the goal is to find a generally applicable reward function.

\begin{table}
    \centering
    \caption{Raw numerical evaluation results ($\mu\pm\sigma$) for the peak performance (PP) and generalization performance (GP) for EUREKA, LEARN-Opt and LEARN-Opt with metrics.}\label{input-ablation-results}%
    \begin{tabular}{p{2.5cm}cC{2.5cm}C{2.5cm}}\toprule
        \textbf{Framework}& \textbf{}& \textbf{PP}& \textbf{GP}\\
        \midrule
        \multicolumn{4}{l}{\textbf{Cartpole}}\\
        \midrule
        \multicolumn{2}{l}{Baseline}& -	& $(4.80\pm0.97)\mathrm{e}{-3}$\\
        \midrule
        \multirow{2}{*}{EUREKA} & Best& $3.82\mathrm{e}{-3}$            & $(4.52\pm0.65)\mathrm{e}{-3}$\\
                                & Mean& $(8.22\pm0.78)\mathrm{e}{-3}$   & $(5.02\pm0.56)\mathrm{e}{-3}$\\
        \midrule
        \multirow{2}{*}{LEARN-Opt}  & Best& $3.52\mathrm{e}{-3}$            & $(3.40\pm0.57)\mathrm{e}{-3}$\\
                                    & Mean& $(6.66\pm0.31)\mathrm{e}{-3}$   & $(5.59\pm1.2)\mathrm{e}{-3}$\\
        \midrule
        LEARN-Opt       & Best& $3.39\mathrm{e}{-3}$            & $(3.38\pm0.52)\mathrm{e}{-3}$\\
        with metrics    & Mean& $(3.55\pm0.01)\mathrm{e}{-3}$   & $(6.88\pm3.4)\mathrm{e}{-3}$\\
        \midrule
        \multicolumn{4}{l}{\textbf{Quadcopter}}\\ 
        \midrule
        \multicolumn{2}{l}{Baseline}& -	& $0.42\pm0.08$\\
        \midrule
        \multirow{2}{*}{EUREKA} & Best& $0.31$              & $0.34\pm0.06$\\
                                & Mean& $0.45\pm0.21$       & $0.54\pm0.30$\\
        \midrule
        \multirow{2}{*}{LEARN-Opt}  & Best& $0.29$            & $0.39\pm0.06$\\
                                    & Mean& $0.64\pm0.54$   & $0.75\pm0.49$\\
        \midrule
        LEARN-Opt       & Best& $0.26$          & $0.37\pm0.04$\\
        with metrics    & Mean& $0.35\pm0.07$   & $0.46\pm0.089$\\
        \midrule
        \multicolumn{4}{l}{\textbf{Ant}}\\ 
        \midrule
        \multicolumn{2}{l}{Baseline}& -	& $7.04\pm0.71$\\
        \midrule
        \multirow{2}{*}{EUREKA} & Best& $7.39$            & $7.82\pm0.37$\\
                                & Mean& $5.28\pm2.23$   & $5.55\pm2.17$\\
        \midrule
        \multirow{2}{*}{LEARN-Opt}  & Best& $8.33$            & $8.07\pm0.42$\\
                                    & Mean& $6.27\pm2.40$   & $5.30\pm2.50$\\
        \midrule
        LEARN-Opt       & Best& $8.77$          & $7.49\pm0.48$\\
        with metrics    & Mean& $8.12\pm0.42$   & $7.01\pm0.41$\\
        \midrule
        \multicolumn{4}{l}{\textbf{Humanoid}}\\ 
        \midrule
        \multicolumn{2}{l}{Baseline}& -	& $5.79\pm0.49$\\
        \midrule
        \multirow{2}{*}{EUREKA} & Best& $5.61$            & $5.90\pm0.60$\\
                                & Mean& $4.14\pm1.26$   & $3.99\pm1.19$\\
        \midrule
        \multirow{2}{*}{LEARN-Opt}  & Best& $5.71$            & $5.42\pm0.23$\\
                                    & Mean& $4.93\pm0.67$   & $4.46\pm0.52$\\
        \midrule
        LEARN-Opt       & Best& $5.70$          & $5.14\pm0.76$\\
        with metrics    & Mean& $5.33\pm0.32$   & $4.53\pm0.33$\\
        \midrule
        \multicolumn{4}{l}{\textbf{Franka-Cabinet}}\\ 
        \midrule
        \multicolumn{2}{l}{Baseline}& -	& $0.36\pm0.35$\\
        \midrule
        \multirow{2}{*}{EUREKA} & Best& $0.15$            & $0.26\pm0.25$\\
                                & Mean& $0.01\pm0.04$   & $0.07\pm0.10$\\
        \midrule
        \multirow{2}{*}{LEARN-Opt}  & Best& $0.82$            & $0.52\pm0.35$\\
                                    & Mean& $0.36\pm0.36$   & $0.15\pm0.18$\\
        \midrule
        LEARN-Opt       & Best& $0.84$          & $0.54\pm0.36$\\
        with metrics    & Mean& $0.38\pm0.37$   & $0.16\pm0.22$\\
        \botrule
     \end{tabular}
\end{table}

\subsection{Impact of the number of analyzers and number of generated metrics.} Given that the modular design of the evaluation module in the LEARN-Opt framework allows for the implementation of an ensemble classifier that incorporates different analyzers and metrics, as we described in Subsection~\ref{section:ensemble classifier}, this introduces two new hyperparameters to the system that need to be carefully selected: the number of metrics and the number of analyzers.

To demonstrate their impact during the optimization process, we conducted the following experiment: we randomly selected 150 experiment sets, which represent the numerical results of a set of reward function candidates executed by the RL framework. Then we used the evaluation module to determine the best reward function candidate from the given sets. In parallel, we utilize the same numerical information, along with the metrics defined in Subsection~\ref{evaluation-metrics}, to identify the best candidates. By using both classifications, we evaluate whether the best reward function candidate selected by the framework aligns with the user's best candidate, using pre-defined evaluation metrics, and thereby determine the hyperparameter values that best fit the proposed framework.

We carried out two variants: first, we kept the number of metrics constant while varying the number of analyzers. Second, we set the number of analyzers constant and we varied the number of generated metrics. After running this process across all experiment sets, we used the well-known accuracy metric to analyze the performance and tune these two hyperparameters. We repeated this procedure for the Ant, Humanoid, and Franka-cabinet tasks, then we averaged the final results. 

\begin{figure}
    \centering
    \includegraphics[width=0.95\linewidth]{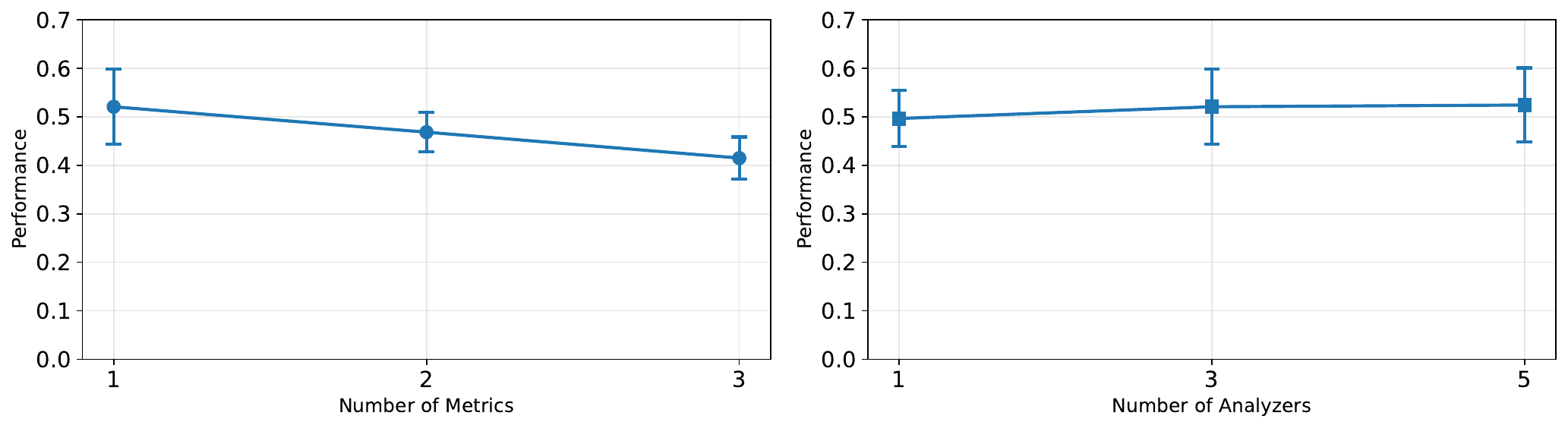}
    \caption{Supporting results to observe the impact of the number of metrics and analyzers. (Left) Performance obtained when varying the number of metrics (analyzers=3). (Right) Performance obtained when varying the number of analyzers (Metrics=1).}
    \label{fig:support-results}
\end{figure}

The results of this experiment are presented in Figure~\ref{fig:support-results}. An initial inspection of this figure suggests that performance decreases as the number of metrics increases, while increasing the number of analyzers enhances performance. However, there is no significant difference when selecting more than three analyzers. Therefore, these results also validate our framework's configuration in the main experiments, where the number of metrics is set to one and the number of analyzers to three to achieve the optimal balance of accuracy and efficiency.

\subsection{LEARN-Opt with different LLM models}\label{secA1}

\begin{figure}[!h]
    \centering
    \includegraphics[width=0.8\linewidth]{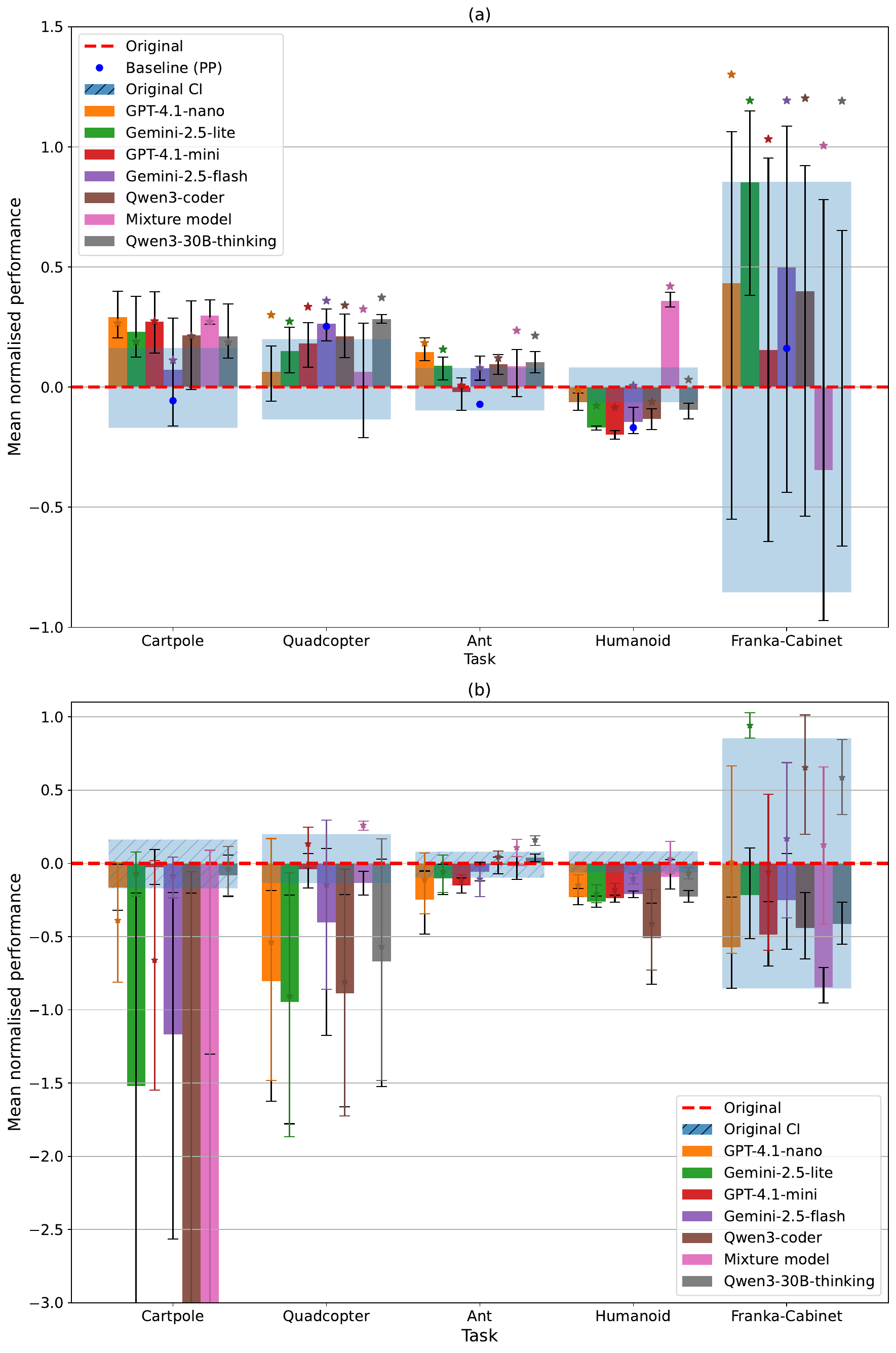}
    \caption{Mean Normalized performance for LEARN-Opt with different LLM models across the five tasks. (a) Best result obtained, (b)Average of all results}
    \label{fig:learnopt_model_ablation}
\end{figure}

We conducted an experiment to observe the impact of using a different LLM model with the LEARN-Opt framework. For this test, we used the following models: \texttt{gpt-4.1-nano}, \texttt{gpt-4.1-mini}, \texttt{gemini-2.5-flash-lite}, \texttt{gemini-2.5-flash}, \texttt{qwen3-coder}, and \texttt{qwen3-30B-thinking}, as well as a mixture model that combines \texttt{gpt-4.1-nano} for the generation module and \texttt{gpt-4.1-mini} for the evaluation module. The results are presented in Figure~\ref{fig:learnopt_model_ablation}.

In general, the results observed in Figure~\ref{fig:learnopt_model_ablation}b reinforce the challenge of finding a good reward function in one-shot, given that almost all models present a negative mean GP. This finding demonstrates that a multi-run methodology is necessary to find high-performing candidates. This fact is also reinforced by observing Figure~\ref{fig:learnopt_model_ablation}a, which presents the best candidates found by the multi-run search. From this figure, we observe that the GP values of the best candidates in almost all models show positive gains on 4 out of 5 tasks, which contrasts with the PP values.

Nevertheless, on the Humanoid task, which failed in our previous experiments, the \textit{mixture model} is the only model that finds a successful candidate. This suggests the proposed framework could also unlock other LLM capabilities for complex, high-dimensional tasks. 

In conclusion, these results affirm that the LEARN-Opt framework is model-agnostic and validate its use for identifying promising candidates within a multi-run schema. In addition, our findings indicate that although good candidates can be found using higher and more complex LLMs, our proposed framework architecture allows the identification of high-performance candidates using low-complexity and smaller LLMs, such as \texttt{gpt-4.1-nano} and \texttt{gemini-2.5-flash-lite}. This demonstrates that LEARN-Opt can also be highly efficient in cost terms, which is another consideration for real-world applications.

\section{Conclusion}
In this paper, we introduce LEARN-Opt, an LLM-based framework that automates the discovery of reward functions in code format from natural language descriptions, reducing human effort when designing reward functions for various control tasks. Compared with state-of-the-art frameworks such as EUREKA, our framework is fully autonomous, enabling the generation and optimization of reward function candidates using natural-language descriptions of the system and the task objective as inputs. In addition, LEARN-Opt addresses EUREKA's limitation for autonomous candidate assessment by building its own evaluation system to assess the different reward functions candidates and select the most promising ones for later refining.

Our experimental evaluation demonstrated that LEARN-Opt's best reward function candidates, found through a multi-run process, achieve par or better performance than EUREKA for most of the tasks, successfully producing reward function candidates that outperform the original handcrafted reward function across a set of evaluated tasks. From these results, we can say that LEARN-Opt achieves comparable performance without relying on pre-defined metrics, thus demonstrating its effectiveness in discovering similar solutions autonomously.

In addition, we demonstrated that the LEARN-Opt's performance can be further enhanced when pre-defined metrics are available. Our ablation study also validated the use of natural language descriptions of the system and the task objective as inputs. Finally, this study also showed that the automated reward design is a high-variance, complex problem, and that the average-case under-performs the original reward function. This finding indicates that a multi-run methodology is necessary in order to find better candidates.

Nevertheless, despite these results, our approach presents some limitations. For example, our framework could not consistently discover a superior reward function for the Humanoid task, suggesting that the actual architecture may struggle to find better candidates as the environment becomes increasingly complex. Moreover, although a multi-run methodology can identify promising candidates, it can be computationally inefficient. Finally, there is room to reduce the overfitting of reward functions to a single seed (although this does not affect the quality of the resulting policy).

Future work could focus on improving search and resource efficiency by adding more agents that can track or detect weaknesses in the reward function via video analysis using VLMs, thereby enhancing the system's capabilities. 

Finally, the LEARN-Opt's modular architecture enhances its flexibility by allowing the use of various specialised LLMs, which can enhance the functionality of the different agent modules within the presented system and enable experimentation with both open- and closed-source solutions without significant modifications.

\bibliography{bibliography.bib}% common bib file

\newpage
\begin{appendices}

% eureka details
\section{EUREKA limitations}\label{secB1}

As detailed in~\cite{ma2023eureka}, EUREKA extracts context concerning the states and the actions the systems use by inspecting the source code from the \texttt{compute\_observations()} function in the IsaacGym simulator, or the \texttt{get\_observations()} function in the last IsaacLab update. While this strategy can effectively capture useful information about the system to generate the reward function candidates, this approach could also introduce a dependency on the implementation of the environment source code. More specifically in scenarios were the needed information comes from internal states that are not easily accessible or directly observable by the users, as in the case of the fitness function construction.

For instance, the IsaacLabEureka release shows a divergence between the observations and the fitness function calculation for the Ant and Humanoid environments as presents Listing~\ref{code:ant_observations} and Listing~\ref{code:eureka_f_ant} respectively. In this case, the corresponding fitness function calculates the mean displacement using \texttt{self.potentials}, a variable that is not present directly in the \texttt{get\_observations()} function. 

Calculating these metrics by using exposed observation variables, for example the \texttt{self.vel\_loc} variable, can be difficult due to internal restrictions of the environment's code as presents the Listing~\ref{code:mine_f_ant}, which restructures the fitness function with the calculation of the mean \textit{x}-velocity. However, this solution utilizes particular internal components of the environment, which, although they work correctly for this case, could present complications if they need to be restructured for another task objective.

\begin{lstlisting}[language=Python, caption=Observation function for Ant and Humanoid environments, label={code:ant_observations}]
def _get_observations(self) -> dict:
        obs = torch.cat(
            (
                self.torso_position[:, 2].view(-1, 1),
                self.vel_loc,
                self.angvel_loc * self.cfg.angular_velocity_scale,
                normalize_angle(self.yaw).unsqueeze(-1),
                normalize_angle(self.roll).unsqueeze(-1),
                normalize_angle(self.angle_to_target).unsqueeze(-1),
                self.up_proj.unsqueeze(-1),
                self.heading_proj.unsqueeze(-1),
                self.dof_pos_scaled,
                self.dof_vel * self.cfg.dof_vel_scale,
                self.actions,
            ),
            dim=-1,
        )
        observations = {"policy": obs}
        return observations
\end{lstlisting}

\begin{lstlisting}[language=Python, caption=Fitness function implemented by EUREKA, label={code:eureka_f_ant}]
(self.potentials - self.prev_potentials).mean()
\end{lstlisting}

\begin{lstlisting}[language=Python, caption=Fitness function restructured to calculate mean x-velocity, label={code:mine_f_ant}]
compute_rot(
    compute_heading_and_up(
        self.robot.data.root_quat_w, 
        self.inv_start_rot, 
        (self.targets - self.robot.data.root_pos_w), 
        self.basis_vec0, 
        self.basis_vec1, 
        2
    )[0], 
    self.robot.data.root_lin_vel_w, 
    self.robot.data.root_ang_vel_w, 
    self.targets, 
    self.robot.data.root_pos_w
)[0][:, 0].mean()
\end{lstlisting}

%=============================================%%
% For submissions to Nature Portfolio Journals %%
% please use the heading ``Extended Data''.   %%
%=============================================%%

%=============================================================%%
% Sample for another appendix section			       %%
%=============================================================%%

% \section{Example of another appendix section}\label{secA2}%
% Appendices may be used for helpful, supporting or essential material that would otherwise 
% clutter, break up or be distracting to the text. Appendices can consist of sections, figures, 
% tables and equations etc.
\section{System descriptions}\label{system_descriptions}

\subsection{Cartpole environment}
\begin{lstlisting}[basicstyle=\fontfamily{\ttdefault}\scriptsize, breaklines=true]
SYSTEM DESCRIPTION
------------------

## Description
A pole is attached by an un-actuated joint to a cart, which moves along a frictionless track. The pendulum is placed upright on the cart and the goal is to balance the pole by applying forces in the left and right direction on the cart to maintain the pole upright, this means when the pole angle is zero.

## Action Space
- Index: 0 - Push cart to the left or right

- Number of Actions: 1

## Observation Space
- Index: 0 - Pole position
- Index: 1 - Pole velocity
- Index: 2 - Cart position
- Index: 3 - Cart velocity

- Number of Observations: 4

## Starting State
All observations are assigned a uniformly random value in (-0.05, 0.05)

## Episode End
1. Termination: Pole Angle is greater than +/-12 degrees
2. Termination: Cart Position is greater than +/-2.4 (center of the cart reaches the edge of the display)
3. Truncation: Episode length is greater than 300
\end{lstlisting}

\subsection{Quadcopter environment}
\begin{lstlisting}[basicstyle=\fontfamily{\ttdefault}\scriptsize, breaklines=true]
SYSTEM DESCRIPTION
------------------

## Description
This system simulates a Crazyflie copter navigating to a goal point using thrust.

## Action Space
- Index: 0 - Thrust
- Index: 1 - Moment X
- Index: 2 - Moment Y
- Index: 3 - Moment Z

- Number of Actions: 4

## Observation Space
- Index: 0 - Linear Velocity X of the copter
- Index: 1 - Linear Velocity Y of the copter
- Index: 2 - Linear Velocity Z of the copter
- Index: 3 - Angular Velocity X of the copter
- Index: 4 - Angular Velocity Y of the copter
- Index: 5 - Angular Velocity Z of the copter
- Index: 6 - Projected Gravity X of the copter
- Index: 7 - Projected Gravity Y of the copter
- Index: 8 - Projected Gravity Z of the copter
- Index: 9 - Distance of the copter to the Desired Position X
- Index: 10 - Distance of the copter to the Desired Position Y
- Index: 11 - Distance of the copter to the Desired Position Z

- Number of Observations: 12

## Starting State
The system starts with a random initial position of the desired goal point.

## Episode End
The episode terminates if the copter dies (falls below 0.1 or above 2.0 on the z-axis).
\end{lstlisting}

\subsection{Ant environment}
\begin{lstlisting}[basicstyle=\fontfamily{\ttdefault}\scriptsize, breaklines=true]
## Description

This environment is based on the one introduced by Schulman, Moritz, Levine, Jordan, and Abbeel in "High-Dimensional Continuous Control Using Generalized Advantage Estimation". The ant is a 3D quadruped robot consisting of a torso (free rotational body) with four legs attached to it, where each leg has two body parts. The goal is to coordinate the four legs to move by applying torque to the eight hinges connecting the two body parts of each leg and the torso (nine body parts and eight hinges).

## Action Space

The action is represented as a `ndarray` with shape `(8,)`, corresponding to the eight degrees of freedom of the robot. An action represents the torques applied at the hinge joints.

- The actions directly influence the joint positions of:
  - Front left leg
  - Front right leg
  - Left back leg
  - Right back leg
  - Front left foot
  - Front right foot
  - Left back foot
  - Right back foot

## Observation Space

List of Observations:

The observation space is a `ndarray` with shape `(36,)` containing:

**Torso Position (z-axis)**: Height of the robot's torso position.

**Locomotion Velocity**: Linear velocity of the robot in its local coordinate space (x, y, z).
   - X locomotion velocity: Forward speed of the robot.
   - Y locomotion velocity: Lateral speed of the robot.
   - Z locomotion velocity: Vertical speed of the robot.

**Angular Velocity**: Rotational speed of the robot in its local coordinate space (x, y, z).
   - X angular velocity: Roll speed of the robot.
   - Y angular velocity: Pitch speed of the robot.
   - Z angular velocity: Yaw speed of the robot.

**Yaw Angle**: Orientation of the robot with respect to the global coordinate system.

**Roll Angle**: Roll orientation of the robot.

**Angle to Target**: Angular difference between the robot's current heading and the target direction (forward direction).

**Up Projection**: Upward vector projection, indicating the robot's orientation with respect to the vertical axis.

**Heading Projection**: Projection of the robot's heading vector in its local coordinate space.

**Joint Positions**: Current positions of the robot's joints, scaled.
   - 'front_left_leg'
   - 'front_right_leg'
   - 'left_back_leg'
   - 'right_back_leg'
   - 'front_left_foot'
   - 'front_right_foot'
   - 'left_back_foot'
   - 'right_back_foot'

**Joint Velocities**: Velocities of the robot's joints, scaled by `dof_vel_scale`.
   - 'front_left_leg'
   - 'front_right_leg'
   - 'left_back_leg'
   - 'right_back_leg'
   - 'front_left_foot'
   - 'front_right_foot'
   - 'left_back_foot'
   - 'right_back_foot'

**Actions**: Actions currently being taken by the agent.
   - 'front_left_leg'
   - 'front_right_leg'
   - 'left_back_leg'
   - 'right_back_leg'
   - 'front_left_foot'
   - 'front_right_foot'
   - 'left_back_foot'
   - 'right_back_foot'

## Starting State

All observations are initialized with a uniform random value, ensuring variability in the initial conditions.
The ant walks forward when goes through the positive x-direction. In the starting state, the robot is standing still with the initial orientation designed to make it face forward in the x-direction.

## Episode End

The episode can conclude under the following conditions:

1. **Termination**: If the robot's torso altitude is below `0.31`.
2. **Truncation**: The maximum episode length is exceeded (set at 15 seconds).
\end{lstlisting}

\subsection{Humanoid environment}
\begin{lstlisting}[basicstyle=\fontfamily{\ttdefault}\scriptsize, breaklines=true]
## Description

This environment is based on the environment introduced by Tassa, Erez and Todorov in "Synthesis and stabilization of complex behaviors through online trajectory optimization". The 3D bipedal robot is designed to simulate a human. It has a torso (abdomen) with a pair of legs and arms, and a pair of tendons connecting the hips to the knees. The legs each consist of three body parts (thigh, shin, foot), and the arms consist of two body parts (upper arm, forearm).

The humanoid walks forward when goes through the positive x-direction. In the starting state, the robot is standing still with the initial orientation designed to make it face forward in the x-direction.

## Action Space

The action space is a 21-dimensional continuous space representing the torque applied to various joints of the humanoid robot. An action represents the torques applied at the hinge joints.

Joint:
   - Lower waist 0
   - Lower waist 1
   - Right upper arm 0
   - Right upper arm 2
   - Left upper arm 0
   - Left upper arm 2
   - Pelvis
   - Right lower arm
   - Left lower arm
   - Right thigh 0
   - Right thigh 1
   - Right thigh 2
   - Left thigh 0
   - Left thigh 1
   - Left thigh 2
   - Right shin
   - Left shin
   - Right foot 0
   - Right foot 1
   - Left foot 0
   - Left foot 1

## Observation Space

List of Observations:

The observation space is a `ndarray` with shape `(36,)`. The observation space consists of the following parts (in order):

**Torso Position (z-axis)**: Height of the robot's torso position.

**Locomotion Velocity**: Linear velocity of the robot in its local coordinate space (x, y, z).
   - X locomotion velocity: Forward speed of the robot.
   - Y locomotion velocity: Lateral speed of the robot.
   - Z locomotion velocity: Vertical speed of the robot.

**Angular Velocity**: Rotational speed of the robot in its local coordinate space (x, y, z).
   - X angular velocity: Roll speed of the robot.
   - Y angular velocity: Pitch speed of the robot.
   - Z angular velocity: Yaw speed of the robot.

**Yaw Angle**: Orientation of the robot with respect to the global coordinate system.

**Roll Angle**: Roll orientation of the robot.

**Angle to Target**: Angular difference between the robot's current heading and the forward direction.

**Up Projection**: Upward vector projection, indicating the robot's orientation with respect to the vertical axis.

**Heading Projection**: Projection of the robot's heading vector in its local coordinate space.

**Joint Positions**: Current positions of the robot's joints, scaled.
   - Lower waist 0
   - Lower waist 1
   - Right upper arm 0
   - Right upper arm 2
   - Left upper arm 0
   - Left upper arm 2
   - Pelvis
   - Right lower arm
   - Left lower arm
   - Right thigh 0
   - Right thigh 1
   - Right thigh 2
   - Left thigh 0
   - Left thigh 1
   - Left thigh 2
   - Right shin
   - Left shin
   - Right foot 0
   - Right foot 1
   - Left foot 0
   - Left foot 1

**Joint Velocities**: Velocities of the robot's joints, scaled by `dof_vel_scale`.
   - Lower waist 0
   - Lower waist 1
   - Right upper arm 0
   - Right upper arm 2
   - Left upper arm 0
   - Left upper arm 2
   - Pelvis
   - Right lower arm
   - Left lower arm
   - Right thigh 0
   - Right thigh 1
   - Right thigh 2
   - Left thigh 0
   - Left thigh 1
   - Left thigh 2
   - Right shin
   - Left shin
   - Right foot 0
   - Right foot 1
   - Left foot 0
   - Left foot 1

**Actions**: Actions currently being taken by the agent.
   - Lower waist 0
   - Lower waist 1
   - Right upper arm 0
   - Right upper arm 2
   - Left upper arm 0
   - Left upper arm 2
   - Pelvis
   - Right lower arm
   - Left lower arm
   - Right thigh 0
   - Right thigh 1
   - Right thigh 2
   - Left thigh 0
   - Left thigh 1
   - Left thigh 2
   - Right shin
   - Left shin
   - Right foot 0
   - Right foot 1
   - Left foot 0
   - Left foot 1


1. **Termination:**
   - If the robot's torso height falls below `0.8`.
\end{lstlisting}

\subsection{Franka environment}
\begin{lstlisting}[basicstyle=\fontfamily{\ttdefault}\scriptsize, breaklines=true]
SYSTEM DESCRIPTION
------------------

## Description
The `Franka Cabinet Env` is based on the 9 degrees of freedom Franka robot. It simulates a robotic environment where a Franka robotic arm interacts with a cabinet placed in front of the robot. The robot's objective is to move the robot's end effector towards a desired position and grasp a cabinet drawer. The robot needs to manipulate the cabinet drawer by applying joint movements and control its end effector and perform actions such as opening or closing the drawer. The cabinet drawer (door) is considered open when the position of the cabinet's door joint is greater than 0.35. 

## Action Space
The action is a `ndarray` with shape `(num_envs, 9)` representing the motor actions for the Franka robot's joints. Each action is internally used to incrementally update the desired joint positions of the robot arm. Then, the environment uses position control where the low-level controller moves the robot joints to the desired positions based on the actions provided by the agent. Each action corresponds to a joint in the robot arm, including the gripper joints. The action values are normalized between -1.0 and 1.0, where:

### Joint Names
- `panda_joint1`: Increment/decrement shoulder pan position 
- `panda_joint2`: Increment/decrement shoulder lift position
- `panda_joint3`: Increment/decrement upper arm rotation
- `panda_joint4`: Increment/decrement elbow flex
- `panda_joint5`: Increment/decrement forearm rotation
- `panda_joint6`: Increment/decrement wrist flex
- `panda_joint7`: Increment/decrement wrist rotation
- `panda_finger_joint1`: Increment/decrement left gripper finger opening
- `panda_finger_joint2`: Increment/decrement right gripper finger opening

**Note**: Actions are clamped within the joint position limits and scaled by the control parameters defined in the environment configuration.

## Observation Space
The observation is a `ndarray` with shape `(num_envs, 23)` representing the state of the environment. The observations include the robot's joint positions, velocities, and the state of the cabinet door. The observations are normalized and scaled to facilitate learning.

List of Observations:

1. **Joint Positions**: 
   - **Description**: Scaled positions of the robot's joints, normalized between -1.0 and 1.0.
   - **Dimension**: `(num_envs, 9)`

   - `Shoulder pan (base)`: hinge joint angle value
   - `Shoulder lift`: hinge joint angle value
   - `Upper arm rotation`: hinge joint angle value
   - `Elbow flex`: hinge joint angle value
   - `Forearm rotation`: hinge joint angle value
   - `Wrist flex`: hinge joint angle value
   - `Wrist rotation`: hinge joint angle value
   - `Left gripper finger`: slide joint translation value
   - `Right gripper finger`: slide joint translation value

2. **Joint Velocities**: 
   - **Description**: Scaled robot joint velocities.
   - **Dimension**: `(num_envs, 9)`

   - `Shoulder pan (base)`: joint angular velocity
   - `Shoulder lift`: joint angular velocity
   - `Upper arm rotation`: joint angular velocity
   - `Elbow flex`: joint angular velocity
   - `Forearm rotation`: joint angular velocity
   - `Wrist flex`: joint angular velocity
   - `Wrist rotation`: joint angular velocity
   - `Left gripper finger`: slide joint linear velocity
   - `Right gripper finger`: slide joint linear velocity

3. **Target Position (Delta)**: 
   - **Description**: The distance difference in position between the robot's grasp position and the target drawer grasp position for the three axes. Vector from gripper to target.
   - **Dimension**: `(num_envs, 3)`

4. **Cabinet Door Position**: 
   - **Description**: The position of the cabinet's door joint, indicating how much is open or closed. The position of the drawer's joint (e.g., how far the drawer is opened).
   - **Dimension**: `(num_envs, 1)`

5. **Cabinet Door Velocity**: 
   - **Description**: The velocity of the cabinet's door joint, providing insight into the speed at which it's opening or closing.
   - **Dimension**: `(num_envs, 1)`

- Number of Observations: 23

## Starting State
All observations are initialized based on the robot's default joint positions with a small random perturbation applied to encourage exploration.

## Episode End
The episode ends if any one of the following occurs:

1. Termination: If the cabinet door joint is greater than 0.35, it means the drawer is opened, so the task is considered complete (terminated). But if this value achieves 0.39, the environment is reset.
   - **Success**: The task is considered successful when the cabinet door joint position exceeds 0.35, indicating the drawer has been successfully opened.
   - **Failure**: If the cabinet door joint position does not exceed 0.35 after a maximum of 500 steps, the task is considered failed.
   - **Reset**: The environment is reset if the cabinet door joint position exceeds 0.39, indicating a need to reset the state for the next episode.
2. Truncation: Episode length exceeds the maximum specified duration of 500 steps (forced to end due to time limit, not necessarily task success/failure).
\end{lstlisting}

\section{RL algorithm hyperparameters}\label{app:ppo_parameters}

% \wc{Network parameters}

Values taken from the RL-Games configuration from the IsaacLab repository\footnote{https://github.com/isaac-sim/IsaacLab}.

\begin{table}[h]
    \caption{A2C hyperparameters used by the RL-Games framework}
    \centering
    \begin{tabular}{cc}
    % ... table content
    \toprule
    \textbf{Parameter} & \textbf{Value}\\
    \midrule
        Discount factor & 0.99\\
        Tau & 0.95\\
        Entropy coefficient & 0\\
        Minibatch size & 1024\\
        KL threshold & 0.016\\
        Grad. norm. & 1\\
        Entropy coefficient & 0 \\
        Sequence length & 4\\
        Bound loss coefficient & 0.0001 \\
    \botrule
    \end{tabular}
\end{table}

\begin{table}[h]
    \caption{Hyperparameters for the environments used by the RL-Games framework}
    \centering
    \begin{tabular}{cccccc}
    % ... table content
    \toprule
    \textbf{Parameter} & \textbf{Cartpole}& \textbf{Quadcopter}& \textbf{Ant}& \textbf{Humanoid}& \textbf{Franka}\\
    \midrule
        MLP units & [32, 32] & [64, 64]& [256, 128, 64]& [400, 200, 100]& [256, 128, 64]\\
        Horizon length & 32& 24& 16& 32& 16\\
        Minibatch size & 16384 & 24576 & 32768 & 32768 & 8192\\
        Mini-epochs & 8& 5& 4& 5& 8\\
        Critic coef. & 4 & 2 & 2 & 4 & 4\\
        Learning rate & 5e-4 & 5e-4& 3e-4& 5e-4& 5e-4\\
    \botrule
    \end{tabular}
\end{table}

\end{appendices}

\end{document}